\pdfoutput=1

\documentclass[11pt]{article}

\usepackage[]{acl}

\usepackage{times}
\usepackage{latexsym}

\usepackage[T1]{fontenc}

\usepackage[utf8]{inputenc}

\usepackage{microtype}

\usepackage{url}

\usepackage{inconsolata}
\usepackage{multirow}
\usepackage{CJKutf8}
\usepackage{xcolor}
\usepackage{makecell}
\usepackage{colortbl}
\usepackage{booktabs}
\usepackage{color}
\usepackage[normalem]{ulem}
\usepackage{graphicx}
\usepackage{CJKutf8}
\usepackage{bbding}
\usepackage{float}
\usepackage{comment}
\usepackage{tabularray}

\title{Defining and Detecting Vulnerability in Human Evaluation Guidelines: \\A Preliminary Study Towards Reliable NLG Evaluation}

\author{Jie Ruan$^{*}$, Wenqing Wang$^{*}$, Xiaojun Wan\\
Peking University\\
\{ruanjie,wangwenqing\}@stu.pku.edu.cn\\ wanxiaojun@pku.edu.cn}

\begin{document}

\maketitle
\def\thefootnote{*}\footnotetext{Equal contribution. 
}\def\thefootnote{\arabic{footnote}}
\begin{abstract}
Human evaluation serves as the gold standard for assessing the quality of Natural Language Generation (NLG) systems. 
Nevertheless, the evaluation guideline, as a pivotal element ensuring reliable and reproducible human assessment, has received limited attention.
Our investigation revealed that only 29.84\% of recent papers involving human evaluation at top conferences release their evaluation guidelines, with vulnerabilities identified in 77.09\% of these guidelines.
Unreliable evaluation guidelines can yield inaccurate assessment outcomes, potentially impeding the advancement of NLG in the right direction. 
To address these challenges, we take an initial step towards reliable evaluation guidelines and
propose the first human evaluation guideline dataset 
by collecting annotations of guidelines extracted from existing papers as well as generated via Large Language Models (LLMs). 
We then introduce a taxonomy of eight vulnerabilities 
and formulate a principle for composing  evaluation guidelines. 
Furthermore, a method for detecting guideline vulnerabilities has been explored using LLMs, and we offer a set of recommendations to enhance reliability in human evaluation. 
The annotated human evaluation guideline dataset and code for the vulnerability detection method are publicly available online.\footnote{\url{https://github.com/EnablerRx/GuidelineVulnDetect}}
\end{abstract}

\section{Introduction}

Natural Language Generation (NLG) has found extensive applications across diverse domains. Nevertheless, evaluating the quality of generated outputs has posed a longstanding and formidable challenge due to the inherent diversity of expressions capable of conveying the same meaning \cite{howcroft2020twenty,zhou2022deconstructing}. This abundance of possible variations complicates the development of automated evaluation methods \cite{novikova2017we,reiter2009investigation}, thus necessitating the reliance on human evaluation as the gold standard and regarding it as a more reliable evaluation method in NLG \cite{celikyilmaz2020evaluation,gatt2018survey,gkatzia2015snapshot,Mellish_Dale_1998,van-der-lee-etal-2018-evaluating}.

However, the evaluation guidelines, which play a crucial role in ensuring the reliability of human evaluation, have not received adequate emphasis. The transparency issues inherent in human evaluation guidelines raise concerns regarding the validity and reproducibility of the evaluation results \cite{schoch-etal-2020-problem}.
To investigate this issue, we conducted a study based on 3,233 papers that we crawled
from ACL, EMNLP, and NAACL conferences in the last three years. Surprisingly, 
we indicate that only 29.84\% of the papers involving human evaluation release their human evaluation guidelines.
Human evaluation guidelines are crucially important for ensuring that human assessments are conducted reliably. However, when papers fail to release the evaluation guidelines, there is no guarantee of the reliability and reproducibility of their evaluation results.
Moreover, our analysis of the guidelines released by these papers uncovered a significant concern: a striking 77.09\% of the released guidelines exhibited noticeable vulnerabilities\footnote{
In this paper, "vulnerability" carries the same meaning as "defect", indicating issues within evaluation guidelines that could potentially result in unreliable evaluation outcomes.}, which could potentially have a detrimental impact on the correctness of human evaluation outcomes  \cite{schoch-etal-2020-problem}.


The ultimate goal of establishing reliable human evaluation guidelines comprises several essential steps, which include detecting potential vulnerabilities in the guidelines, identifying specific vulnerability types, 
marking the precise segments with vulnerabilities,
providing modification suggestions, and finally correcting identified vulnerabilities in the guidelines.
 In this paper, we conduct a preliminary study on defining and detecting vulnerabilities in human evaluation guidelines, marking an initial step towards reliable guidelines.
Specifically, we first constructed a human evaluation guideline dataset by collecting annotations of guidelines extracted from existing papers as well as generated via LLMs.
Based on the analysis of the guidelines, we identified eight main categories of vulnerabilities 
including Ethical Issues, Unconscious Bias, Ambiguous Definition, Unclear Rating, Edge Cases, Prior Knowledge, Inflexible Instructions, and Others. The 
guidelines with vulnerabilities can result in issues such as annotators being unclear about task requirements, misunderstanding specific scoring 
standards, or erroneously directing annotators to assign higher scores to particular systems, which leads to incorrect and irreproducible evaluation results.
To detect these vulnerabilities, we explored several prompt strategies to evoke the capability of current LLMs in vulnerability detection for human evaluation guidelines, and recommend an LLM-based method employing Chain of Thought (CoT) strategies.


The main contribution of this paper is as follows:
1) We are the first to study vulnerabilities in human evaluation guidelines and release the first human evaluation guideline dataset with annotated vulnerabilities for advancing reliable human evaluation; 
2) We analyze the existing human evaluation guidelines and 
introduce a taxonomy of eight vulnerabilities for evaluation guidelines;  Furthermore, we establish a principle for writing a reliable human evaluation guideline; 
3) We explore an LLM-based method for detecting guideline vulnerabilities. We recommend employing this method to assess the reliability of the guidelines before conducting human evaluations; 
4) We present a set of recommendations designed to elevate the reliability of human evaluation by offering guidance on writing robust guidelines and identifying potential vulnerabilities.

\section{Raw Guideline Dataset}

Due to the lack of existing work related to human evaluation guideline assessment, we construct the first 
human evaluation guideline dataset through two methods: extracting from existing papers and generating from GPT3.5, referred to as the \textbf{authentic guidelines} and \textbf{synthetic guidelines}, respectively. Note that we collect and analyze synthetic guidelines because LLMs has been proved as powerful tools for synthetic data generation \citep{agrawal2022qameleon, liu-etal-2022-wanli, bitton2023q2d}. 

\subsection{Authentic Guidelines}
The construction of authentic guidelines involves a three-step process: First, we crawled papers\footnote{We crawled https://paperswithcode.com, an open resource website which ensures our access to the guideline data once they are publicly available. } on ACL, EMNLP, and NAACL conferences from 2020 to 2022 and obtained 3,233 raw data. Then, we filter the papers using two groups of keywords, and narrow down the paper set to 319.  
\textit{Human evaluation} and \textit{manual assessment} constitute the first group, using them to focus solely on the 677 papers related to the evaluation tasks, while
\textit{guideline}, \textit{instruction}, \textit{questionnaire}, \textit{interface}, and \textit{screenshot} are employed as keywords to identify papers potentially containing
guideline sections.
Finally, we manually filtered out papers specifically related to NLG tasks and extract 227 guidelines from 
ACL (111),
EMNLP (62) and NAACL (54). Any guidelines presented as figures or charts were converted into textual formats. More Details of the collected data can be found in Appendix \ref{sec:authentic_guideline_details}.

\subsection{Synthetic Guidelines}
Constructing effective prompts for language models to perform NLP tasks is currently a highly researched topic \citep{schick-schutze-2021-shot, le-scao-rush-2021-many, tam-etal-2021-improving, logan-iv-etal-2022-cutting, reynolds2021prompt}. 
Inspired by \citet{mishra-etal-2022-reframing}, 
we design 5 prompts to guide GPT-3.5-Turbo in generating diverse guidelines, including raw prompt, raw prompt with evaluation aspects, structured prompt, structured prompt with evaluation aspects and structured prompt with evaluation aspects and constraints, as shown in Appendix \ref{sec:prompt_template}. For each prompt, we expand the dataset by incorporating 12 NLG tasks and 2 evaluation settings, along with alternately utilizing the keywords \textit{instruction} and \textit{guideline}. Consequently, we generated a total of 48 guidelines for each prompt (12 tasks × 2 settings × 2 keywords).


\subsection{Data Statistics}
Finally, we obtained 227 authentic guidelines extracted from existing papers, alongside 239\footnote{A piece of synthetic guideline that doesn't belong to the evaluation task has been filtered out.} synthetic guidelines generated by GPT-3.5-Turbo with average lengths of 247.64 words and 237.05 words, respectively. In total, our dataset comprises 466 human evaluation guidelines. It is worth noting that out of the 677 papers related to human evaluation, 
only 202 (29.84\%) of them openly released their evaluation guidelines after 
considering cases where multiple guidelines were included in a single paper, indicating the insufficient attention given to the evaluation guidelines.

\section{Guideline Vulnerability Annotation}

\subsection{Taxonomy of Guideline Vulnerability}  
\label{sec:Vulnerability_Definition}
\begin{table*}[t]\small
\resizebox{\linewidth}{!}{
\centering

\begin{tabular}{|p{17cm}|}
\hline
{\cellcolor[rgb]{0.9,0.9,0.9}}\\
{\cellcolor[rgb]{0.9,0.9,0.9}}\large\textbf{\centerline{Guideline for Opinion Summarization Quality Evaluation}}                                                                                                                                                                                                                                                                                                                                                                                                                                                                                                                                                                                                                                                                                                                                                                                                                                                                                                                                                                                                                                                                                                                                                                                                                                                                                                                                                                                                                                                                                                                                                                                                                                                                                                                                                                                                                                                                                                                                                                                                                                                                                                                                                                                                                                                                                                                                                                                                                                                                                                                                                                                                                                                                                                                                                                                                                                                                                                                                                                                                                                                                                                                                                                                                                                                                                                                                            {\cellcolor[rgb]{0.9,0.9,0.9}} \\
\multicolumn{1}{|p{17cm}|}{\begin{tabular}[c]{@{}p{17cm}@{}}
\\
Thank you for participating in this opinion summarization quality evaluation task!\\Opinion summarization is the task of automatically generating summaries for a set of reviews about a specific target. In this task, we focus on movie reviews written by users from the Rotten Tomatoes website\textsuperscript{{\textbf{1}}}. You will be presented with one human-written reference summary first along with three system summaries generated by trained neural networks respectively\textsuperscript{{\textbf{2}}}. Please evaluate the quality of opinion summaries\textsuperscript{{\textbf{3}}} with respect to the following four features: (1) Relevance; (2) Consistency; (3) Fluency; and (4) Coherence\textsuperscript{{\textbf{4}}}. You should make comparisons for the summary evaluation and rank the four summaries in the order of the four evaluation aspects\textsuperscript{{\textbf{5}}}, and the evaluation is conducted on the open-source annotation tool Doccano\textsuperscript{{\textbf{6}}}.\\\\\textbf{IMPORTANT:}\\\\\begin{tabular}{@{\labelitemi\hspace{\dimexpr\labelsep+0.5\tabcolsep}}p{16.6cm}@{}}Thoroughly read the guideline and familiarize yourself with the task of opinion summarization quality evaluation. \\Carefully read the source reviews as well as reference and system summaries to grasp the overall content.
\\Evaluate the overall quality of each summary based on the four designated aspects, assign a score to each dimension sentence by sentence and aggregate all the scores of each sentence to perform pairwise comparisons\textsuperscript{{\textbf{7}}}.\\If you encounter any difficulties or have questions during the annotation procedure, refer to the provided guidelines. Alternatively, feel free to contact us via email for further clarification.\end{tabular}\end{tabular}}                                                                                                                                                                                                                                                                                                                                                                                                                                                                                                                                                                                                                                                                                                                                                                                                                                                                                                                                                                                                                                                                                                                                                                                                                                                                                                                                                                                                                                                                                                                                        \\ 
\\
\hline
{\cellcolor[rgb]{0.9,0.9,0.9}}\\
{\cellcolor[rgb]{0.9,0.9,0.9}}\large\textbf{\centerline{Vulnerabilities in Guideline}}                                                                                                                                                                                                                                                                                                                                                                                                                                                                                                                                                                                                                                                                                                                                                                                                                                                                                                                                                                                                                                                                                                                                                                                                                                                                                                                                                                                                                                                                                                                                                                                                                                                                                                                                                                                                                                                                                                                                                                                                                                                                                                                                                                                                                                                                                                                                                                                                                                                                                                                                                                                                                                                                                                                                                                                                                                                                                                                                                                                                                                                                                                                                                                                                                                                                                                                                                                      {\cellcolor[rgb]{0.9,0.9,0.9}}\\
\\
\multicolumn{1}{|p{17cm}|}{\begin{tabular}[c]{@{}p{17cm}@{}}{\textbf{1.~Ethical Issues: }}guiding in this manner disregards the personal privacy of the commenters as it fails to specify whether the comments are anonymous or obtained with user consent. An improved guideline should address ethical concerns such as “All anonymized reviews have been previously collected with user consent and have been stripped of personally identifiable information.”\\{\textbf{2.~Unconscious Bias:~}}guiding in this manner specifies the sequence of the summaries, leading evaluators to have a biased perception of the reference as superior in quality. An improved guideline should be more neutral such as “You will be presented with four summaries in a random order, including one reference summary and three system summaries generated by trained neural networks."\\{\textbf{3.~Ambiguous Definition:}}~guiding in this manner fails to clarify whether the task is to evaluate four summaries based on the source review or to evaluate three system-generated summaries based on the reference. An improved guideline should provide a more explicit task definition such as “Please evaluate the quality of both the reference and three system-generated opinion summaries given the corresponding source review."\\{\textbf{4.~Unclear Rating:}}~guiding in this manner lacks a clear explanation of the evaluation aspect, which leads to multiple interpretations for different evaluators, resulting in inconsistent ratings. Given that this task involves pairwise comparison, an improved guideline doesn't have to provide a rating scale, yet it should 
explicit the evaluation criteria such as: “(1) Relevance: measures how well the summary captures the key points of the source review; (2) Consistency: measures whether the facts in the summary are consistent with the facts in the source review; (3) Fluency: measures the quality of individual sentences, are they well-written and grammatically correct; (4) Coherence: measures the quality of all sentences collectively, to the fit together and sound naturally."\\{{\textbf{5.~Edge Cases:}}}~guiding in this manner fails to provide directions for addressing edge cases where both summaries have equal quality. An improved guideline should comprehensively consider exceptional situations such as: “In the case of two summaries are of equal quality, place them side by side in the same ranking."\\{\textbf{6.~Prior Knowledge:}}~guiding in this manner assumes evaluators have annotation experience without explaining how to use the professional tool Doccano. An improved guideline should offer training or detailed explanations for professional tools and principles 
such as “The evaluation is conducted on the open-source annotation tool Doccan, and subsequently, training will be provided on how to use it for annotation. If you are interested, you can visit this website in advance for more information: https://doccano.github.io/doccano."\\{\textbf{7.~Inflexible Instructions:}}~guiding in this manner makes the task unnecessarily complex by aggregating individual sentences for overall quality evaluation. Furthermore, it doesn't align with certain aspects, such as coherence, which require an evaluation that considers all sentences collectively.
An improved guideline should be more flexible and reasonable such as “Evaluate the overall quality of each summary and make comparisons based on the four designated aspects."\end{tabular}}  \\
\\
\hline
\end{tabular}
}
\caption{
 An illustration of the taxonomy on guideline vulnerability types. 
}
\label{tab1}
\end{table*}

We define a typology consisting of eight guideline vulnerabilities by analyzing the guidelines extracted from existing papers and generated by LLMs. An illustration for each type is shown in Table \ref{tab1}, which is designed for illustrative purposes and does not originate from the actual dataset. More examples can be found in Appendix \ref{sec:anno_guideline}.


\noindent\textbf{Ethical Issues} \citep{mieskes-2017-quantitative}: instructions do not consider potential ethical implications related to the evaluation process, like privacy, cultural sensitivity, accessibility, or the potential misuse of the evaluation results.

\noindent\textbf{Unconscious Bias} \citep{schoch-etal-2020-problem}: 
instructions 
unconsciously favors or disadvantages certain results.

 \noindent\textbf{Ambiguous Definition} \citep{jurgens-2014-analysis}: instructions for task definition
 are unclear, vague, or imprecise that can be interpreted in multiple ways.

\noindent\textbf{Unclear Rating} \citep{amidei-etal-2019-use}: 
 instructions that lack standardized criteria for evaluating aspects or 
 definition of each point on a rating scale, resulting in potential inconsistency in ratings.

\noindent\textbf{Edge Cases} \citep{ruggeri2023definition}: instructions do not specify how to handle edge cases or exceptional situations that don't neatly fit into the usual categories or criteria.

\noindent\textbf{Prior Knowledge} \citep{sun-etal-2020-investigating}: instructions assume that evaluators have certain background knowledge or familiarity with a specific subject matter, tool, or principle.

\noindent\textbf{Inflexible Instructions}: instructions are unnecessarily complex or rigid, making it hard for evaluators to follow and incapable of adjusting to variations in data or task requirements, which contradicts \citet{sabou-etal-2014-corpus}'s conclusion that a simpler instruction tends to yield better results.

Finally, we add the additional type \textbf{Others} to ensure the completeness of the typology.
This covers any vulnerabilities that do not fall into the above categories.


\subsection{Data Annotation}
\label{sec:annotation}
We recruit four college students who possess English qualification certificates.
Firstly, they were provided with an annotation guideline, which can be found in Appendix \ref{sec:anno_guideline}. 
Each evaluator went through a training process (details in Appendix \ref{sec:anno_details}) to enhance their understanding 
in the annotation process. Before annotation, we also designed a qualification test consisting of 10 guidelines, only annotators who passed the test were considered qualified and allowed to continue annotation. To ensure the annotation quality, we divided the dataset into batches and assigned the specific number of daily tasks to each annotator. Upon receiving the daily annotations, we reviewed the results and required the specific annotator to reannotate the batch of data assigned for that day if there was a low accuracy (less than 80\%).

In the annotation interface, the authentic guidelines and synthetic guidelines are randomly presented on the left side so as to prevent bias, while the eight vulnerability types are displayed on the right.
Annotators were instructed to assign the specific vulnerability types based on the predefined typology, or indicate “None" for guidelines where the vulnerability type is absent. Each sample was annotated by two distinct annotators and a third annotator made the final decision if they are in disagreement. 

We utilized Cohen's kappa \citep{cohen1960coefficient}  to measure the inter-annotator agreement and computed on a per-label basis so as to  gain label-specific insights.
Ultimately, we calculated the mean values across all labels to assess the overall agreement.
The annotation process lasted approximately two weeks, culminating in a substantial inter-annotator agreement of Cohen’s kappa with $\kappa$=0.722 on authentic guidelines and $\kappa$=0.737 on the synthetic guidelines. More annotation details can be found in Appendix \ref{sec:anno_details}.


%

\subsection{Annotation Result}
\begin{figure}[htbp]
    \centering
    \includegraphics[width=\linewidth]{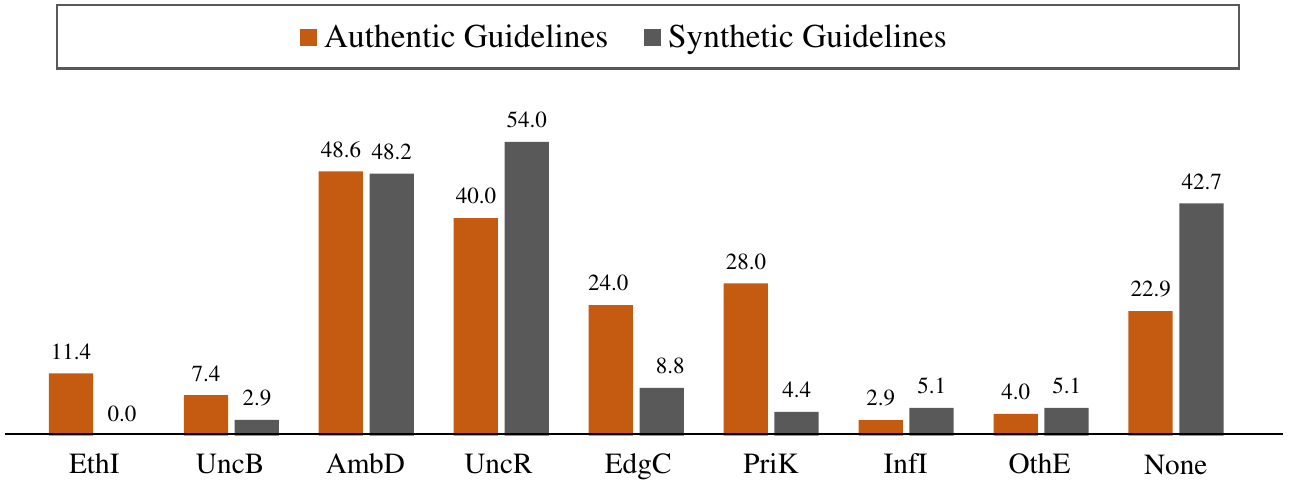}
    \caption{Distributions of vulnerability types on authentic and synthetic guidelines with EthI, UncB, AmbD, UncR, EdgC, PriK, InfI, OthE refers to Ethical Issues, Unconscious Bias, Ambiguous Definition, Unclear Rating, Edge Cases, Prior Knowledge, Inflexible Instructions and Others respectively. “None” means the guideline has no vulnerability at all. The ratio calculation is achieved by taking the number of guidelines that include a particular category and dividing it by the total count of guidelines.
    }
    \label{Fig1}
    \end{figure} 
Figure \ref{Fig1} reports the annotation results on both authentic and synthetic guidelines. While LLMs have shown impressive results in various generation tasks, its current capabilities to generate reliable 
evaluation guidelines is limited, with vulnerabilities over 50\%.
We also report the results of five prompts in Appendix \ref{sec:prompt_template}, indicating that structured instructions incorporating evaluation aspects exhibit the lowest vulnerability ratio.
What is worth noting is that the quality of the authentic guidelines extracted from existing papers is much poorer, and the vulnerability ratio is 77.09\%, which undermines the reliability of evaluation tasks. This aligns with the conclusions of \citet{sabou-etal-2014-corpus}, who demonstrated that the crowdsourcing community still lacks a set of best-practice guidelines, resulting in low-quality annotations. We make a call for future researchers to be aware of the issue and emphasize the need for thorough refinement and investigation to develop a robust guideline.

Regarding the vulnerability type, both the authentic and synthetic guidelines are in a similar distribution that Ambiguous Definition and Unclear Rating occur most frequently. 
The vulnerability "Others" appears in scenarios such as when guidelines generated by LLMs are incomplete.
Table \ref{Figure2} shows the authentic guideline for the machine-in-the-loop writing of image caption task extracted from \citet{padmakumar-he-2022-machine}. The guideline lacks the definition of the machine-in-the-loop writing task and fails to specify the evaluation criterion, leaving uncertainty about the annotation process.
As a result, the reliability and validity of the evaluation process will be compromised. Apart from the two types, authentic guidelines exhibit more vulnerabilities of bias, prior knowledge and ethical issues. Additionally, authentic guidelines are more likely to suffer from neglecting edge cases, whereas the LLM is more prone to generate excessively rigid and complex guidelines, resulting in more vulnerabilities of Inflexible Instructions. 
As such, resorting to the LLMs to fill in the gaps proves to be a promising approach. 
However, it is important to note that the current LLM can only generate preliminary drafts of guidelines and needs more effective strategies to enhance reasoning ability for improving the reliability of guidelines. A future direction is to enhance LLM's reasoning ability to improve its capability in generating reliable guidelines.

\begin{table}[t]
\small
\centering
\begin{tabular}{|p{7.4cm}|}
\hline
\vspace{0.05cm}
\textbf{Instructions for crowdworkers evaluating
the captions}
\\$\bullet$ Choose the better (more descriptive and/or
figurative) caption for the image.
\\$\bullet$ A better caption is your subjective judgement,
the rubrics to make the choice are that the
caption is descriptive and/or figurative in its
interpretation of the image (Refer the examples for further clarification).
\\$\bullet$ The explanation asked is supposed to be very
brief. A single word of if you like it for being
descriptive or interpretive will do.
\\$\bullet$ Relevance of the caption to the image is your subjective choice whether the caption appropriately represents what is in the image and is
not just a catchy piece of text unrelated to the
image.
\\$\bullet$ A caption that you deem irrelevant should
never be the better caption, unless both are
irrelevant.
\\\\
\hline
\end{tabular}
\caption{
Authentic human evaluation guideline extracted from \citet{padmakumar-he-2022-machine} with vulnerabilities of Ambiguous Definition and Unclear Rating.
}
\label{Figure2}
\end{table}
\section{Experiments}
In this section, we investigate utilizing LLMs to detect the specific vulnerability types in each evaluation guideline, which is taken as a multi-label vulnerability type classification task.


\subsection{Large Language Models}
We perform our experiments utilizing both open-source and closed-source LLMs. 
For open-source models, we fine-tuned LLaMA-7B, an efficient and popular foundation language model with LoRA\footnote{https://github.com/Lightning-AI/lit-llama}. 
Additionally, we also experimented with Flan-T5-XXL\footnote{https://huggingface.co/google/flan-t5-xxl}, Flan-Alpaca-L\footnote{https://huggingface.co/declare-lab/flan-alpaca-large}, and Falcon-7B\footnote{https://huggingface.co/tiiuae/falcon-7b}, respectively. For closed-source models, we select two widely accessible large language models: TEXT-DAVINCI-003\footnote{https://platform.openai.com/docs/models/gpt-3-5} and GPT-3.5-turbo\footnote{https://platform.openai.com/docs/models/gpt-3-5}.
TEXT-DAVINCI-003 is developed using a combination of supervised instruction tuning and Reinforcement Learning from Human Feedback methodologies.
GPT-3.5-Turbo is an enhanced version of the GPT-3 language model with instruction fine-tuning.

\subsection{Prompting Strategies}
Our exploration involves designing prompts for both zero-shot and few-shot scenarios, encompassing four distinct prompt templates (“Basic”, “VDesc”, “CoT-Basic” and “CoT-VDesc”) under each scenario, thus yielding a total of eight prompts. 
Basic prompt offers only the name of the vulnerability type, whereas VDesc prompt expands on this by including definition for each type. Additionally, we investigate the Chain-of-Thought (CoT) prompting technique on both prompt templates. Detailed prompting design and the full prompts are detailed in Appendix \ref{sec:Prompt_Template}.

\subsection{Baselines}
\label{baseline}
We further implement and finetune three Transformer-based classifiers as baselines: BERT \citep{devlin-etal-2019-bert} along with its successors XLNet \citep{yang2019xlnet} and ALBERT \citep{lan2019albert}, which have shown excellent performance on classification tasks. They are all deep pretrained models that first encodes a guideline into vector space by capturing contextual information bidirectionally and then outputs the probability for each label independently. We finetune all the models on the base version and the hyper-parameters can be found in Appendix \ref{sec:hyper-para}.

\begin{table*}[t]
\centering
\small
\renewcommand{\arraystretch}{1.1}
\resizebox{\linewidth}{!}{
\begin{tabular}{p{7.5cm}|p{1.8cm}|p{5.5cm}}
\hline
\toprule
\multicolumn{1}{c}{{\cellcolor[rgb]{0.9,0.9,0.9}}\textbf{Human Evaluation   Guideline}}           & \multicolumn{1}{c}{{\cellcolor[rgb]{0.9,0.9,0.9}}\textbf{LLM}} & \multicolumn{1}{c}{{\cellcolor[rgb]{0.9,0.9,0.9}}\textbf{LLM Answer}}\\  \hline
\multirow{7}{*}{\begin{tabular}[c]{@{}p{7.5cm}@{}}Read the following   pairs of texts (source and candidate) and provide ratings between 0 to 100 based on the degree of similarity in \underline{\textbf{meaning}} and preservation of \underline{\textbf{grammar}} for each candidate text. Drag or click on the appropriate portion of the slider to provide a rating for each candidate text. Please refer to the scoring example before starting the first   task.\\ \underline{\textbf{Note:}} There will be a candidate text in each HIT which contains several random words appearing out of context unrelated to the text. Please make sure to score this candidate text in each HIT with \textbf{10-20} points. Apart from this, please refer to the table below for   scoring the candidate text. You must provide a rating for all examples to   proceed. \\\textbf{Category: Meaning}   \\ Mild (1 sent): deduct \textbf{5-10} pts e.g., wrong pronoun     \\ Significant (1 sent): deduct \textbf{30} pts e.g., relevant words but diff   meaning\\ Mild (2 sent): deduct \textbf{10-20} pts\\ Significant (2 sent): deduct \textbf{60-70} pts      \\ \textbf{Category: Grammaticality}      \\ Mild (1 sent): deduct \textbf{5-10} pts e.g., missing conjunction\\ Significant (1 sent): deduct \textbf{25} pts e.g., incomplete sentence\\ Mild (2 sent): deduct \textbf{10-20} pts\\ Significant (2 sent): deduct \textbf{50-60} pts\end{tabular}}  
&  \textcolor[rgb]{0,0.502,0}{\textbf{GOLD}}                                                      & \textcolor{red}{Ambiguous Definition, Unclear Rating} \\ \cline{2-3}                                                                                         & \begin{tabular}[c]{@{}p{1.8cm}@{}}\textcolor[rgb]{0,0.502,0}{\textbf{TEXT-}}\\ \textcolor[rgb]{0,0.502,0}{\textbf{DAVINCI-003}}\end{tabular} & 
\begin{tabular}[c]{@{}p{5.5cm}@{}}REASONING: The guideline does   not provide a clear definition of the task, making it difficult for   evaluators to interpret the instructions. It also does not provide a detailed   explanation of the rating aspects, leading to multiple interpretations for different   evaluators, causing "Unclear Rating". So the answer is   "Ambiguous Definition" and "Unclear Rating".\\ \textcolor{red}{ANSWER: Ambiguous Definition, Unclear Rating}~~\CheckmarkBold\end{tabular} 
\\ \cline{2-3}                                                                                         & \textcolor[rgb]{0,0.502,0}{\textbf{GPT-3.5-Turbo}}                                               & 
\begin{tabular}[c]{@{}p{5.5cm}@{}}REASONING: The guideline does   not provide clear instructions on how to handle edge cases where the   candidate text contains random words unrelated to the text. So the answer is   "Edge Cases"\\ ANSWER: Edge Cases~~\XSolidBrush\end{tabular} 
\\ \cline{2-3}                     & \textcolor[rgb]{0,0.502,0}{\textbf{LLaMA}}                                                       & Mild, Significant~~\XSolidBrush 
\\ \cline{2-3}
&   \begin{tabular}[c]{@{}p{2cm}@{}}\textcolor[rgb]{0,0.502,0}{\textbf{Flan}}\end{tabular} &                                                       \begin{tabular}[c]{@{}p{5.5cm}@{}}Let’s think step by step: REASONING: The first task is to score ~~\XSolidBrush\end{tabular} \\ \cline{2-3} 
&   \textcolor[rgb]{0,0.502,0}{\textbf{Flan-Alpaca}}                                                      & Categoryeaningequal to 12 pts e ~~\XSolidBrush \\ \cline{2-3} 
&   \textcolor[rgb]{0,0.502,0}{\textbf{Falcon}}                                                      & None ~~\XSolidBrush \\ 
\cline{1-3}
\toprule
\end{tabular}
}
\caption{Example answers of vulnerability types detected (generated) by LLMs with few-shot CoT prompting for a human evaluation guideline in the test set.}
\label{tab4}
\end{table*}

\subsection{Data Splits} 
We initially divide the dataset into five parts, with four parts designated for training (80\%) and one for testing (20\%). The training set is used for supervised fine-tuning of pretrained baselines and is subsequently divided into train/validation sets in a 4:1 ratio. Further, each of these five parts is used as an individual testing set, while the remaining four parts serve as training sets. As such, we evaluate the performance of the baselines and LLMs across the entire dataset, treating each part as a test set in rotation, so as to mitigate random fluctuations due to the relatively small size of the dataset and obtain a more accurate performance estimate.

\subsection{Evaluation Metrics} 


Following \citet{chen2017ensemble}, we adopt \textbf{macro-Precision(macro-P)}, \textbf{macro-Recall (macro-R)}, and \textbf{macro-F1} scores \citep{vechtomova-2009-book}, which assess the overall performance of a classifier by taking the average of Precision, Recall, and F1-scores across all individual labels for each class (including ``None''). 
Considering the unequal proportions of different vulnerability types in the dataset, as shown in Figure \ref{Fig1}, macro metrics can provide a more balanced view of the model's performance across all classes, as opposed to micro-averaging \citep{vechtomova-2009-book}, which gives more weight to the larger classes. 
Furthermore, we follow \citet{ganda2018survey} and utilize \textbf{Accuracy (ACC)} to assess the average accuracy of all individual types.
We also follow \citet{wu2017unified} and use the \textbf{instance-AUC (AUC)} metric.
\textbf{Hamming Loss} \citep{schapire1998improved} is also incorporated, which evaluates the fraction of misclassified instance-label pairs, accounting for both missed relevant labels and predicted irrelevant labels.

\section{Results and Analysis}

\subsection{Qualitative Analysis}

We first show the case study of a sample from the test set in Table \ref{tab4}, in which the authentic guideline is drawn from \citet{kim-etal-2021-bisect} and suffers from vulnerabilities of Ambiguous Definition and Unclear Rating. The answers are generated by LLMs under few-shot CoT prompting.
We can find that TEXT-DAVINCI-003 not only generates completely correct answers, but also narrows down the scope of vulnerabilities in its reasoning, facilitating the correction of identified vulnerabilities in the guidelines.
Nevertheless, GPT-3.5-Turbo appears to have misconstrued the definition of the Edge Cases, 
since the handling of cases ``where the candidate text contains random words unrelated to the text'' has already been provided. 
The four open-source models, on the other hand, don't generate the answer as instructed. Instead, LLaMA extracts keywords directly as the output, Flan and Flan-Alpaca yields nonsensical results,  and Falcon consistently outputs ``None'' for all of the data, revealing the inefficiency of open-source models in vulnerability detection. 
We speculate the reason is due to the limited training data and the excessive length of the instructions.


\subsection{Quantitative Analysis}

\begin{table*}[htbp]
\centering
\renewcommand{\arraystretch}{1.1}
\resizebox{\linewidth}{!}{
\begin{tabular}{p{1.5cm}p{1.5cm}p{0.7cm}p{0.7cm}p{0.7cm}p{0.7cm}p{0.7cm}p{0.7cm}p{0.73cm}p{0.73cm}p{0.73cm}p{0.7cm}p{0.7cm}p{0.7cm}p{0.7cm}p{0.7cm}p{0.7cm}p{0.7cm}p{0.7cm}p{0.7cm}}
\hline
\toprule
\multicolumn{1}{c}{\multirow{2}{*}{\textbf{\large Model}}} & \multicolumn{1}{c|}{\multirow{2}{*}{\textbf{\large Prompt}}} & \multicolumn{3}{c|}{\textbf{\large Macro-P}} & \multicolumn{3}{c|}{\textbf{\large Macro-R}} & \multicolumn{3}{c|}{\textbf{\large Macro-F1}} & \multicolumn{3}{c|}{\textbf{\large ACC}} & \multicolumn{3}{c|}{\textbf{\large AUC}} & \multicolumn{3}{c}{\textbf{\large Hamming Loss↓}} \\
\multicolumn{1}{c}{} & \multicolumn{1}{c|}{} & Aut & Syn & \multicolumn{1}{c|}{All} & Aut & Syn & \multicolumn{1}{c|}{All} & Aut & Syn & \multicolumn{1}{c|}{All} & Aut & Syn & \multicolumn{1}{c|}{All} & Aut & Syn & \multicolumn{1}{c|}{All} & Aut & Syn & All \\  \cmidrule(r){1-20}
\multicolumn{20}{c}{{\cellcolor[rgb]{0.9,0.9,0.9}}\textbf{LLMs zero-shot}} \\  \cmidrule(r){1-20}
\multirow{4}{*}{\begin{tabular}[c]{@{}l@{}}\large TEXT-\\ \large DAVINCI-\\ \large 003\end{tabular}} & \multicolumn{1}{l|}{Basic} & 0.43 & 0.49 & \multicolumn{1}{c|}{0.47} & 0.52 & 0.49 & \multicolumn{1}{c|}{0.50} & 0.41 & 0.46 & \multicolumn{1}{c|}{0.44} & 0.69 & 0.72 & \multicolumn{1}{c|}{0.71} & 0.42 & 0.53 & \multicolumn{1}{c|}{0.49} & 0.31 & 0.28 & 0.29 \\
 & \multicolumn{1}{l|}{Vdesc} & 0.43 & 0.47 & \multicolumn{1}{c|}{0.46} & 0.45 & 0.52 & \multicolumn{1}{c|}{0.49} & 0.44 & 0.49 & \multicolumn{1}{c|}{0.47} & 0.76 & 0.81 & \multicolumn{1}{c|}{0.79} & 0.45 & 0.53 & \multicolumn{1}{c|}{0.50} & 0.21 & 0.14 & \textbf{0.17} \\
 & \multicolumn{1}{l|}{CoT-Basic} & 0.49 & 0.49 & \multicolumn{1}{c|}{0.49} & 0.52 & 0.51 & \multicolumn{1}{c|}{0.51} & 0.36 & 0.41 & \multicolumn{1}{c|}{0.39} & 0.54 & 0.59 & \multicolumn{1}{c|}{0.57} & 0.46 & 0.57 & \multicolumn{1}{c|}{0.53} & 0.46 & 0.41 & 0.43 \\
 & \multicolumn{1}{l|}{CoT-Vdesc} & 0.46 & 0.52 & \multicolumn{1}{c|}{0.50} & 0.47 & 0.54 & \multicolumn{1}{c|}{0.51} & 0.39 & 0.46 & \multicolumn{1}{c|}{0.43} & 0.60 & 0.66 & \multicolumn{1}{c|}{0.64} & 0.45 & 0.60 & \multicolumn{1}{c|}{0.54} & 0.40 & 0.34 & 0.36 \\ \cmidrule(r){1-20}
\multirow{4}{*}{\begin{tabular}[c]{@{}l@{}}\large GPT-3.5-\\ \large Turbo\end{tabular}} & \multicolumn{1}{l|}{Basic} & 0.53 & 0.48 & \multicolumn{1}{c|}{0.50} & 0.58 & 0.53 & \multicolumn{1}{c|}{0.55} & 0.48 & 0.41 & \multicolumn{1}{c|}{0.44} & 0.64 & 0.61 & \multicolumn{1}{c|}{0.63} & 0.60 & 0.39 & \multicolumn{1}{c|}{0.55} & 0.36 & 0.51 & 0.37 \\
 & \multicolumn{1}{l|}{Vdesc} & 0.53 & 0.53 & \multicolumn{1}{c|}{0.53} & \textbf{0.61} & 0.55 & \multicolumn{1}{c|}{0.58} & 0.44 & 0.48 & \multicolumn{1}{c|}{0.46} & 0.56 & 0.69 & \multicolumn{1}{c|}{0.64} & 0.63 & 0.60 & \multicolumn{1}{c|}{0.61} & 0.44 & 0.31 & 0.36 \\
 & \multicolumn{1}{l|}{CoT-Basic} & 0.53 & 0.52 & \multicolumn{1}{c|}{0.53} & 0.58 & 0.55 & \multicolumn{1}{c|}{0.56} & 0.50 & 0.52 & \multicolumn{1}{c|}{0.51} & 0.67 & 0.76 & \multicolumn{1}{c|}{0.72} & 0.59 & 0.59 & \multicolumn{1}{c|}{0.59} & 0.33 & 0.24 & 0.28 \\
 & \multicolumn{1}{l|}{CoT-Vdesc} & 0.51 & 0.53 & \multicolumn{1}{c|}{0.53} & 0.57 & 0.55 & \multicolumn{1}{c|}{0.56} & 0.49 & 0.53 & \multicolumn{1}{c|}{0.51} & 0.69 & 0.81 & \multicolumn{1}{c|}{0.76} & 0.55 & 0.58 & \multicolumn{1}{c|}{0.57} & 0.31 & 0.19 & 0.24 \\  \cmidrule(r){1-20}
\multicolumn{20}{c}{{\cellcolor[rgb]{0.9,0.9,0.9}}\textbf{LLMs few-shot}} \\  \cmidrule(r){1-20}
\multirow{4}{*}{\begin{tabular}[c]{@{}l@{}}\large TEXT-\\ \large DAVINCI-\\ \large 003\end{tabular}} & \multicolumn{1}{l|}{Basic} & 0.51 & 0.55 & \multicolumn{1}{c|}{0.53} & 0.58 & 0.51 & \multicolumn{1}{c|}{0.54} & 0.51 & 0.50 & \multicolumn{1}{c|}{0.50} & 0.78 & 0.76 & \multicolumn{1}{c|}{0.77} & 0.54 & 0.54 & \multicolumn{1}{c|}{0.54} & 0.22 & 0.24 & 0.23 \\
 & \multicolumn{1}{l|}{Vdesc} & 0.48 & 0.56 & \multicolumn{1}{c|}{0.53} & 0.52 & 0.55 & \multicolumn{1}{c|}{0.54} & 0.43 & 0.52 & \multicolumn{1}{c|}{0.49} & 0.73 & 0.79 & \multicolumn{1}{c|}{0.77} & 0.50 & 0.61 & \multicolumn{1}{c|}{0.57} & 0.27 & 0.21 & 0.23 \\
 & \multicolumn{1}{l|}{CoT-Basic} & \textbf{0.54} & 0.47 & \multicolumn{1}{c|}{0.50} & 0.57 & 0.46 & \multicolumn{1}{c|}{0.51} & 0.53 & 0.46 & \multicolumn{1}{c|}{0.49} & 0.79 & 0.80 & \multicolumn{1}{c|}{0.79} & 0.65 & 0.59 & \multicolumn{1}{c|}{0.62} & 0.21 & 0.20 & 0.21 \\
 & \multicolumn{1}{l|}{CoT-Vdesc} & \textbf{0.54} & 0.68 & \multicolumn{1}{c|}{\textbf{0.61}} & 0.58 & {0.61} & \multicolumn{1}{c|}{\textbf{0.60}} & \textbf{0.54} & {0.62} & \multicolumn{1}{c|}{\textbf{0.58}} & \textbf{0.80} & \textbf{0.87} & \multicolumn{1}{c|}{\textbf{0.83}} & \textbf{0.67} & \textbf{0.69} & \multicolumn{1}{c|}{\textbf{0.68}} & \textbf{0.20} & \textbf{0.13} & \textbf{0.17} \\  \cmidrule(r){1-20}
\multirow{4}{*}{\begin{tabular}[c]{@{}l@{}}\large GPT-3.5-\\ \large Turbo\end{tabular}} & \multicolumn{1}{l|}{Basic} & 0.44 & 0.60 & \multicolumn{1}{c|}{0.51} & 0.48 & 0.68 & \multicolumn{1}{c|}{0.58} & 0.46 & 0.56 & \multicolumn{1}{c|}{0.51} & 0.71 & 0.85 & \multicolumn{1}{c|}{0.78} & 0.48 & 0.60 & \multicolumn{1}{c|}{0.54} & 0.29 & 0.15 & 0.22 \\
 & \multicolumn{1}{l|}{Vdesc} & 0.45 & 0.60 & \multicolumn{1}{c|}{0.52} & 0.48 & \textbf{0.73} & \multicolumn{1}{c|}{\textbf{0.60}} & 0.46 & 0.58 & \multicolumn{1}{c|}{0.52} & 0.71 & 0.85 & \multicolumn{1}{c|}{0.78} & 0.49 & 0.61 & \multicolumn{1}{c|}{0.55} & 0.29 & 0.15 & 0.22 \\
 & \multicolumn{1}{l|}{CoT-Basic} & 0.46 & 0.55 & \multicolumn{1}{c|}{0.51} & 0.49 & 0.57 & \multicolumn{1}{c|}{0.54} & 0.46 & 0.54 & \multicolumn{1}{c|}{0.51} & 0.74 & 0.85 & \multicolumn{1}{c|}{0.79} & 0.50 & 0.61 & \multicolumn{1}{c|}{0.56} & 0.26 & 0.15 & 0.21 \\
 & \multicolumn{1}{l|}{CoT-Vdesc} & 0.44 & \textbf{0.71} & \multicolumn{1}{c|}{{0.60}} & 0.50 & 0.68 & \multicolumn{1}{c|}{\textbf{0.60}} & 0.46 & \textbf{0.67} & \multicolumn{1}{c|}{\textbf{0.58}} & 0.75 & \textbf{0.87} & \multicolumn{1}{c|}{{0.80}} & 0.50 & 0.66 & \multicolumn{1}{c|}{0.59} & 0.24 & \textbf{0.13} & {0.20} \\  \cmidrule(r){1-20}
\multicolumn{20}{c}{{\cellcolor[rgb]{0.9,0.9,0.9}}\textbf{Baseline}} \\  \cmidrule(r){1-20}
\large BERT & \multicolumn{1}{l|}{} & 0.48 & 0.50 & \multicolumn{1}{c|}{0.49} & 0.48 & 0.50 & \multicolumn{1}{c|}{0.49} & 0.48 & 0.47 & \multicolumn{1}{c|}{0.48} & 0.75 & 0.81 & \multicolumn{1}{c|}{0.79} & 0.52 & 0.53 & \multicolumn{1}{c|}{0.53} & 0.25 & 0.19 & 0.21 \\
\large XLNet & \multicolumn{1}{c|}{} & 0.45 & 0.48 & \multicolumn{1}{c|}{0.47} & 0.48 & 0.49 & \multicolumn{1}{c|}{0.49} & 0.48 & 0.47 & \multicolumn{1}{c|}{0.47} & 0.76 & 0.80 & \multicolumn{1}{c|}{0.79} & 0.56 & 0.52 & \multicolumn{1}{c|}{0.53} & 0.24 & 0.20 & 0.21 \\
\large ALBERT & \multicolumn{1}{c|}{} & 0.45 & 0.49 & \multicolumn{1}{c|}{0.47} & 0.48 & 0.50 & \multicolumn{1}{c|}{0.49} & 0.45 & 0.48 & \multicolumn{1}{c|}{0.46} & 0.79 & 0.80 & \multicolumn{1}{c|}{0.80} & 0.45 & 0.61 & \multicolumn{1}{c|}{0.51} & \textbf{0.20} & 0.21 & 0.20 \\  \hline
\toprule
\end{tabular}} 
\caption{Guideline vulnerability detection results on ``Authentic Guidelines (Aut)'', ``Synthetic Guidelines (Syn)'' and the whole dataset (All). Upper, middle and lower parts show results of LLMs under zero-shot and few-shot scenarios as well as baseline models, respectively. The best values of each column are bolded. ↓ indicates that the lower value indicates the better performance. }
\label{tab:tab6}
\end{table*}

Given that the results generated by open-source LLMs are invalid, as demonstrated in Table \ref{tab4}, quantitative evaluation becomes unfeasible.
Therefore, we focus on GPT models and pre-trained baselines for quantitative analysis.
Table \ref{tab:tab6} shows the experiment results for guideline vulnerability detection on both authentic and synthetic guidelines along with the entire dataset.
We also report the results of each vulnerability type in Appendix \ref{sec:results_of_each_type}.
We first explored the effects of different prompt strategies, including Basic, Vdesc, and the use of CoT. Subsequently, we explored the detection performance of LLMs in zero-shot and few-shot settings. Additionally, we investigated the performance of different LLMs, namely TEXT-DAVINCI-003, GPT-3.5-Turbo, and pre-trained models including BERT, XLNet, and ALBERT. Finally, we analyzed the varying performance between authentic guidelines and synthetic guidelines. Through this exploration of different prompt strategies, models, and settings, we conclude that TEXT-DAVINCI-003 demonstrates superior performance with few-shot prompting and CoT strategies. Our analysis of experimental results exploring different prompt strategies, models, and settings is based on a comprehensive consideration of all evaluation metrics. When drawing conclusions from specific metrics, we specify the particular metrics that serve as the basis for our conclusions.

Regarding the Basic and VDesc prompt templates, they exhibit comparable capabilities. The reason is that the incorporation of vulnerability descriptions might potentially disrupt the reasoning process of LLMs, although they might provide detailed vulnerability descriptions for LLMs. According to results on All guidelines, we can also find that CoT generally improves model performance in all prompt strategies of zero-shot setting and Vdesc prompt strategy of few-shot setting. The reason why CoT doesn't consistently enhance model performance in the Basic prompt strategy may stem from the insufficiency of vulnerability information provided by the Basic prompt for effective reasoning. For few-shot and zero-shot settings, we can conclude from the results on All guidelines that LLMs generally exhibit enhanced performance in few-shot scenarios.

In the analysis of various LLMs and pretrained models, the experimental results indicate that TEXT-DAVINCI-003 and GPT-3.5-Turbo exhibit comparable performance, consistently outperforming pretrained models across the majority of prompt strategies. However, the pretrained models still serve as robust baselines, showing specific advantages over TEXT-DAVINCI-003 without CoT strategies under zero-shot scenarios. A noteworthy observation is that Recall values generally surpass Precision in most cases, indicating a tendency for the models to classify guidelines as positive, i.e., no vulnerability is detected. Furthermore, observing the results of each vulnerability type (detailed experimental results are shown in Appendix  \ref{sec:results_of_each_type}), it is found that the model's ability to detect different vulnerabilities varies significantly. All these gaps suggest that the models still have room for improvement in guideline vulnerability detection.

We also compare the model's performance on the two categories of guidelines: Authentic Guidelines and Synthetic Guidelines. Experimental results in Table \ref{tab:tab6} indicate that LLMs exhibit a stronger ability to detect vulnerabilities in synthetic guidelines compared to authentic guidelines. Moreover, TEXT-DAVINCI-003 with CoT-Vdesc strategy demonstrates superior detection capabilities in detecting authentic guidelines, while GPT-3.5-Turbo with CoT-Vdesc strategy exhibits enhanced detection proficiency for synthetic guidelines. Overall, the experimental results show TEXT-DAVINCI-003 exhibits superior detection capabilities in detecting all guidelines.

Based on the outcomes of a thorough exploration involving various prompt strategies, models, and settings, our conclusion is that TEXT-DAVINCI-003 demonstrates superior performance with few-shot prompting and CoT strategies. Overall, TEXT-DAVINCI-003 with CoT-Vdesc prompt strategy in the few-shot scenario has the best performance for all guidelines and is recommended as the method for guideline vulnerability detection.

\section{Practical Recommendations}
We summarize the key findings from this work and provide practical recommendations for reliable human evaluation.
\begin{enumerate}
    \item Writing human evaluation guidelines using LLMs. Our research has found that the proportion of vulnerabilities in guidelines generated by LLMs is lower than those written by humans. We suggest directly instructing LLMs about the requirements for evaluation and utilizing them to generate human evaluation guidelines.
    \item Modify the evaluation guideline draft written by LLMs based on the proposed principles for human evaluation guidelines (shown in Appendix \ref{sec:principal_guideline}). We analyze human evaluation guidelines and summarize principles to compose a robust evaluation guidelines. 
    We recommend referencing these principles when crafting the guidelines.
    \item Utilize TEXT-DAVINCI-003 with CoT-VDesc strategy to  identify vulnerabilities. 
    It has been proven to be an efficient, convenient, and cost-effective method, and detecting a guideline only requires approximately \$0.02\footnote{The prompt consists of 909 tokens, with the inclusion of the average length of each guideline(242.21 tokens), multiplied by the cost of TEXT-DAVINCI-003 (\$0.0200 / 1K).
}.

    \item Conduct human evaluation in strict accordance with the human evaluation guidelines.
    \item Publicly release the human evaluation guideline. 
    This can contribute to the transparency of human evaluation.
\end{enumerate}

\section{Related Work}  
\subsection{Vulnerability Detection}
Early explorations in Vulnerability Detection (VD) span rule-based approaches targeting predefined patterns. Subsequent advancements incorporate Machine Learning (ML) and Deep Learning (DL) techniques to predict vulnerabilities automatically in various tasks, including software security detection \cite{li2018vuldeepecker}, smart contract opcodes detection \cite{qian2022multi} and code vulnerability detection \cite{cheshkov2023evaluation}.
Recently, inspired by the outstanding performance of LLM in code-based tasks, \citet{cheshkov2023evaluation} attempted to explore the capability of LLM in addressing code vulnerability detection. 
Vulnerability detection already being implemented across a wide range of tasks with various techniques, yet none of them have been designed to explore the issue of vulnerability detection in human evaluation guidelines.

\subsection{Natural Language Generation Evaluation}
Previous studies have frequently relied on automatic metrics like BLEU \citep{papineni-etal-2002-bleu}, METEOR \citep{banerjee-lavie-2005-meteor}, ROUGE \citep{lin-2004-rouge}, BERT-SCORE \cite{zhang2019bertscore}, MOVER-SCORE \cite{zhao2019moverscore} and BART-SCORE \cite{yuan2021bartscore} to evaluate the quality of generated text, primarily due to their cost-effectiveness, quickness, and repeatability \citep{reiter-belz-2009-investigation}. Nevertheless, these metrics have been criticized for their limited interpretability \citep{van-der-lee-etal-2019-best} and low correlation with human judgements \citep{belz-reiter-2006-comparing,liu-etal-2016-evaluate,reiter-belz-2009-investigation,novikova-etal-2017-need}. Human evaluation is widely recognized as the gold standard for evaluating NLG systems \citep{Mellish_Dale_1998,gkatzia-mahamood-2015-snapshot,van-der-lee-etal-2018-evaluating}. However, it has the potential to be unreliable due to cognitive biases \citep{schoch-etal-2020-problem} and the lack of standardized evaluation methodologies \citep{van-der-lee-etal-2019-best}. 
\citet{shimorina2021human} contributed to transparency in the human evaluation process by documenting it, while \citet{belz2023missing} explored reproducibility in NLP human evaluation. \citet{ruan2024better} proposed CASF to solve the sampling problem in human evaluation. However, there is currently no comprehensive work addressing the reliability of human evaluation guidelines, a pivotal element ensuring reliable and reproducible human assessment.
With the increasing interest in LLMs, recent studies have been conducted to examine their suitability for assessing generation tasks \citep{gao2024llm}, like summarization \citep{Luo_Xie_Ananiadou_2023,gao2023human}, machine translation \citep{Kocmi_Federmann}, etc.
In this work, we focus on both human evaluation and large language model evaluation, which is the first to utilize LLMs for assessing guidelines in human evaluation.

\section{Conclusion}
In this paper, we propose and analyze significant issues in the evaluation guidelines of gold-standard human assessments.
We conduct a preliminary study on defining and detecting vulnerabilities in 
guidelines to advancing reliable human evaluation. By proposing a taxonomy of guideline vulnerabilities, we constructed the first annotated human evaluation guideline dataset. We then explored LLMs with Few-Shot prompting and CoT strategies for automatic vulnerability detection. 
Recommendations include employing LLMs to assist in writing human evaluation guidelines and modifying them based on the proposed principles. Utilizing the 
proposed LLM-based vulnerability detection method is suggested for assessing the reliability of the guidelines.

In future work, we will delve into the precise  annotation of 
spans containing vulnerabilities in guidelines, 
providing correction suggestions, 
automatically correcting identified vulnerabilities in the guidelines, and generating reliable guidelines by AI models. This advancement aims to contribute towards the ultimate goal of establishing dependable gold-standard human evaluation guidelines, thereby enhancing the reliability of NLG assessments.


\section*{Acknowledgements}
This work was supported by Beijing Science and Technology Program (Z231100007423011), National Key R\&D Program of China (2021YFF0901502), National Science Foundation of China (No. 62161160339) and Key Laboratory of Science, Technology and Standard in Press Industry (Key Laboratory of Intelligent Press Media Technology). We thank Professor Ehud Reiter for providing constructive suggestions. We appreciate the anonymous reviewers for their helpful comments. Xiaojun Wan is the corresponding author.

\section*{Limitations}
This study serves as a preliminary exploration towards establishing reliable evaluation guidelines. We proposed and analyzed significant issues in gold-standard human assessments, specifically focusing on identifying vulnerabilities in guidelines. Our preliminary study employed LLMs to detect guideline vulnerabilities and provided recommendations for improving reliability in human evaluation.
However, the ultimate goal of achieving dependable gold-standard human evaluation guidelines requires further investigation. Future work can delve into precise annotation of spans containing vulnerabilities, automatic correction of identified issues, and the generation of reliable guidelines using AI models. These advancements aim to contribute to establishing dependable guidelines, thereby enhancing the reliability of NLG assessments.
It is important to note that due to cost considerations, experiments with the proposed method were not conducted on GPT-4. Implementing the proposed method on GPT-4 may further enhance its effectiveness, a consideration for future research.

\section*{Ethics Statement}
We recruit annotators from a college campus. They are completely free to decide whether or not to participate in our annotation. The payment is 9 dollars per hour,  higher than the local minimum wage. There is no personal information in our collected dataset. The information which may be used to identify the participants is deleted after the annotation.
Moreover, the LLM-generated guidelines may contain toxic language, which can make annotators uncomfortable. We reviewed the data before annotation and found no problematic samples.
We check the licenses of the artifacts used in this study and do not find conflicts. The license of the dataset we will release is CC BY-NC 4.0.


\bibliography{main}

\clearpage
\newpage
\newpage
\appendix

\section{Authentic Guidelines Details}
\label{sec:authentic_guideline_details}
 For the collected data, we focused on work related to human evaluation in NLG tasks. In descending order of frequency, specific tasks include summarization (42), dialogue generation(36), question answering (34), machine translation (26), story generation (20), image captioning (9), etc. These guidelines are collected from high-quality NLP conferences ACL, EMNLP and NAACL over the past three years (2020-2022). Apart from machine translation that cover a range of language pairs like English-French, English-Japanese, Chinese -English, English-German, English-Spanish, and English-Russian, most of the tasks primarily focus on the English language. Additionally, we have gathered information on the reported inter-annotator agreement, revealing a general inverse relationship between the number of identified vulnerabilities and the level of agreement. To illustrate, in the vulnerability-free guideline from \citet{jiang-etal-2020-neural}, Cohen's Kappa can reach a substantial level of 0.807, whereas in the guideline from \citet{roy-etal-2021-identifying}, with three identified vulnerabilities, Cohen's Kappa falls only within the range of 0.50 to 0.64. 
 The list of crawled papers and the guidelines with annotations are released.

\section{Prompts for Synthetic Guideline Generation}
\label{sec:prompt_template}
To explore the LLM's ability in writing human evaluation guidelines and extend the guideline datasets, we utilize different prompt strategies for LLMs to generate diverse human evaluation guidelines.
Table \ref{table3} displays the prompts that were employed for creating synthetic guidelines, which fall into two categories: raw instructions and structured instructions.
Inspired by the sensitivity of language models to the framing of their instructional prompts \citep{mishra-etal-2022-reframing}, we explore the impact of incorporating evaluation aspects and constraints separately, with a total of five prompt variations. For each prompt, we analyze their performance across 12 NLG tasks: summarization, machine translation, dialogue generation, story generation, paraphrase generation, data to text, grammar error correction, text simplification, code generation, code summarization, question generation, and spelling correction, involve two assessment methods: direct assessment and pairwise comparison, and focus on the keywords “guideline” and “instruction”.

The annotation result of each prompt can be found in Table \ref{tab13}. It can be seen that structured instructions, as opposed to raw instructions, generally contain fewer vulnerabilities and both can enhance generation performance after adding evaluation aspects. However, incorporating constraints into the prompt leads to a drop in generation performance, contradicting the findings of \citet{shi2022toward}, who employ a fluency constraint and observed an enhancement in performance.

\begin{table}[h]
\renewcommand{\arraystretch}{1.05}
\centering
\resizebox{\linewidth}{!}{
\begin{tabular}{l|c}
\hline
\toprule
\textbf{Prompt} & \multicolumn{1}{l}{\textbf{\begin{tabular}[c]{@{}l@{}}\small \% Vulnerability\\\small Ratio ↓\end{tabular}}} \\ \cmidrule(r){1-2}
raw & 13.8 \\
raw with aspect & 10.5 \\
structured & 10.9 \\
structured with aspect & 9.6 \\
structured with aspect and constraint & 12.6 \\ \hline
\toprule
\end{tabular}
}
\caption{Annotation results regarding the vulnerability ratio of each prompt variation. The ratio calculation involves dividing the count of synthetic guidelines containing vulnerabilities for a specific prompt by the overall guideline count. ↓ indicates a lower value is preferable.}
\label{tab13}
\end{table}

\begin{table*}[t]\small
\centering
\resizebox{\linewidth}{!}{

\begin{tabular}{p{0.00001cm}p{0.2cm}p{15cm}} 
\hline
\toprule
                                                         &                                                                 & \textit{Five prompt variations for two instructions}                                                                                                                                                                                                                                                                                                                                                                                                                                                                                                                                                                                                                                                                                                    \\ 
\hline
\multirow{2}{*}{\raisebox{-8ex}{\textcolor[RGB]{128,0,255}{\rotatebox{90}{RAW}}}}  & \multirow{2}{*}{\raisebox{20ex}{\textcolor[RGB]{128,0,255}{\rotatebox{90}{INSTRUCTION}}}}   & 
\begin{tabular}[c]{@{}p{15cm} @{}}\vspace{0.0001cm}\textcolor[RGB]{191,128,64}{\textbf{Raw prompt:} }\\Write a human evaluation \textcolor{blue}{\textit{guideline}} for the \textcolor{blue}{\textit{Summarization}} task. The evaluation type is \textcolor{blue}{\textit{Pairwise Comparison}}.\vspace{0.13cm}\end{tabular}                                                                                                                                                                                                                                                                                                                                                                                                                                                                                       \\ 

                                                         &                                                                 & \begin{tabular}[c]{@{}p{15cm} @{}}\vspace{0.001cm}\textbf{\textcolor[RGB]{191,128,64}{Raw prompt with evaluation aspects:}}\\ Write a human evaluation \textcolor{blue}{\textit{guideline}} for the \textit{\textcolor{blue}{Summarization}} task. The evaluation type is \textcolor{blue}{\textit{Pairwise Comparison}}. Evaluate the following aspects: accuracy, coherence, consistency, relevance, fluency, informativeness, coverage, overall. The evaluation scale is 1-5 (1 is poor and 5 is excellent). \vspace{0.1cm}\end{tabular}                                                                                                                                                                                                                               \\ 
\cline{2-3}
\multirow{3}{*}{\raisebox{-22ex}{\textcolor[RGB]{128,0,255}{\rotatebox{90}{STRUCTURED}}}} & \multirow{3}{*}{\raisebox{-22ex}{\textcolor[RGB]{128,0,255}{\rotatebox{90}{INSTRUCTION} }}} & \begin{tabular}[c]{@{}p{15cm} @{}}\vspace{0.0001cm}\textcolor[RGB]{191,128,64}{\textbf{Structured prompt: }}\\Human evaluation task: \textcolor{blue}{\textit{Summarization }}\\Evaluation type: \textcolor{blue}{\textit{Pairwise Comparison }}\\Human evaluation \textcolor{blue}{\textit{guideline}}:\vspace{0.13cm}~\end{tabular}                                                                                                                                                                                                                                                                                                                                                                                                                                              \\ 

                                                         &                                                                 & \begin{tabular}[c]{@{}p{15cm} @{}}\vspace{0.00001cm}\textcolor[RGB]{191,128,64}{\textbf{Structured prompt with evaluation aspects:~ }}~\\Human evaluation task: \textcolor{blue}{\textit{Summarization}}~ \\Evaluation type: \textcolor{blue}{\textit{Pairwise Comparison}}\\Evaluation aspects: accuracy, coherence, consistency, relevance, fluency, informativeness, coverage, overall\\Evaluation scale: 1-5 (1 is poor and 5 is excellent)\\Human evaluation \textcolor{blue}{\textit{guideline}}:\vspace{0.13cm}\end{tabular}                                                                                                                                                                                                                                                \\ 

                                                         &                                                                 & \begin{tabular}[c]{@{}p{15cm} @{}}\vspace{0.00001cm}\textcolor[RGB]{191,128,64}{\textbf{Structured prompt with evaluation aspects and constraints:~ }}~\\Human evaluation task: \textcolor{blue}{\textit{Summarization}}~ \\Evaluation type: \textcolor{blue}{\textit{Pairwise Comparison}}~\\Evaluation aspects: accuracy, coherence, consistency, relevance, fluency, informativeness, coverage, overall~ \\Evaluation scale: 1-5 (1 is poor and 5 is excellent)\\Please be mindful of the following issues and avoid them: definition ambiguity, bias, assuming prior knowledge, insufficient coverage, lack of rating scale, lack of adaptability, and neglecting ethical implications\\Human evaluation \textcolor{blue}{\textit{guideline}}:~ ~\end{tabular}  \\
\bottomrule
\end{tabular}
}
\caption{
Five prompts utilized for generating synthetic guidelines. The highlighted blue portions are employed across 12 NLG tasks, 2 evaluation methods and 2 key words, which are interchanged interchangeably.
}
\label{table3}
\end{table*}

\section{Annotation Guideline}
\label{sec:anno_guideline}
We release our full guideline provided to crowdworker participants for the manual evaluation of the vulnerability detection task in Table \ref{table4}. We advocate for more related works to share their guidelines, aiming to enhance the transparency of human evaluation and thereby contribute to the establishment of a set of best-practice guidelines for the community.

\begin{table*}[t]\small
\centering
\setlength{\extrarowheight}{0pt}
\addtolength{\extrarowheight}{\aboverulesep}
\addtolength{\extrarowheight}{\belowrulesep}
\setlength{\aboverulesep}{0pt}
\setlength{\belowrulesep}{0pt}
\resizebox{\linewidth}{!}{
\begin{tabular}{|p{18cm}|} 
\hline
\multicolumn{1}{|c|}{\cellcolor[rgb]{0.9,0.9,0.9}{\large\textbf{Vulnerability Detection in Human Evaluation Guidelines}}}                                                                                                                                                                                                                                                                                                                                                                                                                                                                                                                                                                                                                                                                                                                                                                                                                                                                                                                                                                                                                                                                                                                                                                                                                                                                                                                                                                                                                                                                                                                                                                                                                                                                                                                                  \\ 
\hline
\textbf{\large\textcolor{red}{Task Overview}}                                                                                                                                                                                                                                                                                                                                                                                                                                                                                                                                                                                                                                                                                                                                                                                                                                                                                                                                                                                                                                                                                                                                                                                                                                                                                                                                                                                                                                                                                                                                                                                                                                                                                                                                                               \\
Thank you for participating in this task! We are currently working on a project focused on crafting robust and reliable guidelines for human evaluation. You will be randomly presented with a human evaluation guideline extracted from existing papers or generated by Large Language Models (LLMs). Your job is to review the provided guidelines and identify potential vulnerabilities within the text. These vulnerabilities should fall into one or more of the eight categories outlined below.\\                                                                                                                                                                                                                                                                                                                                                                                                                                                                                                                                                                                                                                                                                                                                                                                                                                                                                                                                                                                                                                                                                                                                                                                                                                                                                                                          \\ 
\hline
\textbf{\large\textcolor{red}{Defect Types}}                                                                                                                                                                                                                                                                                                                                                                                                                                                                                                                                                                                                                                                                                                                                                                                                                                                                                                                                                                                                                                                                                                                                                                                                                                                                                                                                                                                                                                                                                                                                                                                                                                                                                                                                                                \\
\begin{tabular}[c]
{@{}p{18cm}@{}}\textbf{Ethical Issues: }instructions do not consider potential ethical implications related to the evaluation process, like privacy, cultural sensitivity, accessibility, or the potential misuse of the evaluation results.\\\begin{tabular}{@{\labelitemi\hspace{\dimexpr\labelsep+0.5\tabcolsep}}p{17.5cm}@{}}Ethical Issues: Evaluate the comments on this public social media post for sentiment analysis.\\Improved: Evaluate anonymized comments provided for sentiment analysis. All comments have been previously collected with user consent and have been stripped of personally identifiable information.~ ~ ~~\end{tabular}\end{tabular} \\

\begin{tabular}[c]{@{}p{18cm}@{}}\textbf{Unconscious Bias:} instructions 
unconsciously favors or disadvantages certain results.\\\begin{tabular}{@{\labelitemi\hspace{\dimexpr\labelsep+0.5\tabcolsep}}p{17.5cm}@{}}Unconscious Bias: Evaluate the two systems A and B: How many points do you think system A is higher than system B?\\Improved: Evaluate the two systems A and B based on user satisfaction and score them respectively.~ ~ ~~\end{tabular}\end{tabular}

\\
\begin{tabular}[c]
{@{}p{18cm}@{}}\textbf{Ambiguous Definition:} instructions for task definition
 are unclear, vague, or imprecise that can be interpreted in multiple ways.\\\begin{tabular}{@{\labelitemi\hspace{\dimexpr\labelsep+0.5\tabcolsep}}p{17.5cm}@{}}Ambiguous Definition:\textit{Factual consistency of summarization} is defined as the accuracy and faithfulness of a summary in representing the source.\\Improved: \textit{Factual consistency of summarization} is defined as the accuracy and faithfulness of a summary in representing the source. The \textit{source} here usually has scenarios: the first is the input document and the second is common sense. In our task, we only focus on the first situation, i.e. evaluate the summary as factually inconsistent if it contains extra information of the input document,even though it is true facts.~ ~ ~~\end{tabular}\end{tabular}                                                                                                                                                                                      \\
\begin{tabular}[c]{@{}p{18cm}@{}}\textbf{Unclear Rating:} instructions that lack standardized criteria for evaluating aspects or definition of each point on a rating scale, resulting in potential inconsistency in ratings.\\\begin{tabular}{@{\labelitemi\hspace{\dimexpr\labelsep+0.5\tabcolsep}}p{17.5cm}@{}}Unclear Rating: Rate the quality of the website.\\Improved: Rate the quality of the website based on its design, ease of navigation, and relevance of content on a scale of 1 to 5, where 1 is 'very poor' and 5 is 'excellent'. If the website was generally good but had one major flaw, consider rating it a 3or 4 depending on the severity of the flaw. If the website was poor but had one saving grace, consider rating it a 2 or 3.~ ~ ~~\end{tabular}\end{tabular}                                                                                                                                  \\
\begin{tabular}[c]{@{}p{18cm}@{}}\textbf{Edge Cases:~}instructions do not specify how to handle edge cases or exceptional situations that don't neatly fit into the usual categories or criteria.\\\begin{tabular}{@{\labelitemi\hspace{\dimexpr\labelsep+0.5\tabcolsep}}p{17.5cm}@{}}Edge Cases: Evaluate the factuality error types in the summary, including: Hallucination Error, Entity Error, Particulars Error, Predicate Error, Coreference Error.\\Improved: Evaluate the factuality error types in the summary, including: Hallucination Error, Entity Error, Particulars Error, Predicate Error, Coreference Error. If the summary contains multiple errors, please list them all. If the error does not correspond to any of the above types, evaluate it as “Others”.~ ~ ~~\end{tabular}\end{tabular}                                                                                                                                                                                                                                                                                                                                                                                                                                                                                                                                                                                                                                                                                                                                                                                                                                                                                                                                                                                                                                                                                                                                                                                                                                                                                                                                                                                                                                                                                                                                                                                                                                                                                                                                                                                                                                                                                                                                                                                                                                    \\
\begin{tabular}[c]{@{}p{18cm}@{}}\textbf{Prior knowledge:~}instructions assume that evaluators have certain background knowledge or familiarity with a specific subject matter, tool, or principle.\\\begin{tabular}{@{}p{17.5cm}@{}}{\labelitemi}\hspace{\dimexpr\labelsep+0.5\tabcolsep}Prior knowledge: Evaluate the use of object-oriented programming (OOP) principles in the code.\\{\labelitemi}\hspace{\dimexpr\labelsep+0.5\tabcolsep}Improved: Evaluate the use of object-oriented programming (OOP) principles in the code. Check for the following aspects. If you are unfamiliar with these principles, please refer to https://baldur.gitbook.io/patters-and-best-practices/solid/oop-principles for more information."~ ~~\\\hspace{0.5\leftmargin}{\labelitemii}\hspace{\dimexpr\labelsep+0.5\tabcolsep}Encapsulation:Object properties are hidden, and object properties need to be modified through object methods.\\\hspace{0.5\leftmargin}{\labelitemii}\hspace{\dimexpr\labelsep+0.5\tabcolsep}Inheritance:Subclasses can inherit the properties and methods of the parent class without redefining them.\\\hspace{0.5\leftmargin}{\labelitemii}\hspace{\dimexpr\labelsep+0.5\tabcolsep}Polymorphism: Polymorphism can be divided into static and dynamic. Static means that the same object can have different forms of expression, while dynamic means that a parent type can point to an instance of its subtype, making the subtype respond differently to the same method.\\\hspace{0.5\leftmargin}{\labelitemii}\hspace{\dimexpr\labelsep+0.5\tabcolsep}Abstraction:Abstraction refers to extracting the common attributes and behaviors of a class and storing them in a class, regardless of how the specific behaviors a rerealized.~ ~ ~\end{tabular}\end{tabular}                                                                                                                                                                                                                                                                                                                                                                                                                                                                                                                                                                                                                                                                                                                                                                                                                                                                                                                                                                                                                                                                                                                                                                                                                                                                                                                                                                                                                                                                                                                                                                                                                                                                                                                                                                                                                                                                                                                                  \\
\begin{tabular}[c]{@{}p{18cm}@{}}\textbf{Inflexible Instructions:} instructions are unnecessarily complex or rigid, making it hard for evaluators to follow and incapable of adjusting to variations in data or task requirements.\\\begin{tabular}{@{\labelitemi\hspace{\dimexpr\labelsep+0.5\tabcolsep}}p{17.5cm}@{}}Inflexible Instructions:Evaluate the website's user interface design on a scale of 1 to 10, considering color aesthetics, balance between text and imagery, navigability, font choices,button placements, menu design, adherence to modern design principles, web page loading speed, and responsive design.\\Improved: Evaluate the website's user interface design on a scale of 1 to 10 from the perspectives of aesthetics, navigation and functionality.~ ~ ~~\end{tabular}\end{tabular}                                                                                                                                                                                                                                                                                                                                                                                                                                                                                                                                                                                                                                                                                                                                                                                                                                                                                                                                                                                                                                                                                                                                                                                                                                                                                                                                                                                                                                                                                                                                                                                                                                                                                                                                                                                                                                                                                                                                                                                                                                                                                                            \\
\textbf{Others: }covers any vulnerabilities that do not fall into the above categories.~ ~ ~ ~         \\                                                                                                                                                                  \\
\hline
\end{tabular}
}
\end{table*}                                                                                                                                                                                                                                                                                                                                                                                                                                                                                                                                                                                                                                                                                                                                                                                                                                               \begin{table*}[t]\small
\centering
\setlength{\extrarowheight}{0pt}
\addtolength{\extrarowheight}{\aboverulesep}
\addtolength{\extrarowheight}{\belowrulesep}
\setlength{\aboverulesep}{0pt}
\setlength{\belowrulesep}{0pt}
\resizebox{\linewidth}{!}{
\begin{tabular}{|p{18cm}|} 
\hline
\textit{(Continued from previous page)}\\ 
\hline
\textbf{\large\textcolor{red}{Annotation Procedure}}                                                                                                                                                                                                                                                                                                                                                                                                                                                                                                                                                                                                                                                                                                                                                                                                                                                                                                                                                                                                                                                                                                                                                                                                                                                                                                                                                                                                                                                                                                                                                                                                                                                                                                                                                        \\
\begin{tabular}{@{\labelitemi\hspace{\dimexpr\labelsep+0.5\tabcolsep}}p{17.5cm}@{}}\textbf{Comprehension: }Carefully read through the entire human evaluation guideline in the center of the interface to get a full understanding of the content.\\\textbf{Labeling: }Identify and click all potential vulnerabilities within the guideline according to the eight defined categories: Definition Ambiguity, Bias, Assuming Prior Knowledge, Insufficient Coverage, Lack of Rating Scale, Lack of adaptability, Neglecting Ethical Implications and Others.\\\textbf{Review and Submit: }Repeat this process until the entire guideline has been thoroughly reviewed and all potential vulnerabilities have been identified. Press Enter to save and submit the annotation result.\\\end{tabular}                                                                                                                                                                                                                                                                                                                                                                                                                                                                                                                                                                                                                                                                                                         \\ 
\hline
\textbf{\large\textcolor{red}{Emphasis and Caution}}                                                                                                                                                                                                                                                                                                                                                                                                                                                                                                                                                                                                                                                                                                                                                                                                                                                                                                                                                                                                                                                                                                                                                                                                                                                                                                                                                                                                                                                                                                                                                                                                                                                                                                                                                        \\
\begin{tabular}{@{\labelitemi\hspace{\dimexpr\labelsep+0.5\tabcolsep}}p{17.5cm}@{}}\textbf{Edge Cases: }Please note that a single guideline may contain more than one type of defect. In such scenarios,ensure to label all the appropriate defect types. If the defect does not fit any of the seven specific categories, classify it as "Others" and provide a brief explanation.\\\textbf{Daily Annotation Requirement:} The guideline for this task will be provided in batches. You are required to annotate a setof 30 items each day. Please submit the daily annotation results before 24:00(midnight) of that day.\\\textbf{Quality Assurance:} Each day, we will conduct a random inspection of the annotated data. If the accuracy rate falls below 80\%, you will be required to re-annotate the data for that day. Please maintain high quality in your annotations.\\\textbf{Support and Reference:} If you encounter any confusion regarding professional knowledge or context while performing this task, please feel free to reach out to us for clarification.You may also refer to Wikipedia or other reliable sources to gain further understanding.\\\textbf{Feedback Mechanism:} Wehave set up a discussion board on the interface, where you can directly submit your queries, concerns, or suggestions through the button “click to comment on document”. This collaborative environment will allow for shared learning and problem-solving.~ ~ ~~\end{tabular}                                                                                                                                                                                                                                                                                                                                               \\ 
\hline
\textbf{\large\textcolor{red}{Examples of Defect Labelling}}                                                                                                                                                                                                                                                                                                                                                                                                                                                                                                                                                                                                                                                                                                                                                                                                                                                                                                                                                                                                                                                                                                                                                                                                                                                                                                                                                                                                                                                                                                                                                                                                                                                                                                                                                \\
\cellcolor[rgb]{0.9,0.9,0.9}{\textbf{\textit{POSITIVE EXAMPLES}}}                                                                                                                                                                                                                                                                                                                                                                                                                                                                                                                                                                                                                                                                                                                                                                                                                                                                                                                                                                                                                                                                                                                                                                                                                                                                                                                                                                                                                                                                                                                                                                                                                                                                                                                          \\
\textbf{Guideline: }Ensure you partake in a comprehensive interrogation of the quantifiable parameters that govern the efficacy of the experimental intervention under scrutiny, taking into account the numerous facets and intricate variables that contribute to the overall outcome, keeping in mind the statistical significance thresholds and the corresponding probability distributions. Your final judgement should be a synthesis of these insights, crystallized into a ranking that encapsulates the overall potency of the intervention in question.                                                                                                                                                                                                                                                                                                                                                                                                                                                                                                                                                                                                                                                                                                                                                                                                                                                                                                                                                                                                                                                                                                                                                                                                                                               \\
\textbf{Label:} Assuming Prior Knowledge, Lack of adaptability, Lack of Rating Scale ~   \CheckmarkBold                                                                                                                                                                                                                                                                                                                                                                                                                                                                                                                                                                                                                                                                                                                                                                                                                                                                                                                                                                                                                                                                                                                                                                                                                                                                                                                                                                                                                                                                                                                                                                                                                                                                                     \\
\begin{tabular}[c]{@{}p{17.5cm}@{}}\textbf{Explanation: }\\Assuming Prior Knowledge: The guideline contains terms and concepts such as"quantifiable parameters," "efficacy of the experimental intervention," "statistical significance thresholds," and"probability distributions". These terms assume the annotators process a prior knowledge in statistics or experimental design, which might make them struggle to understand and apply these instructions correctly without specific training or explanation of the professional background.\\~Lack of adaptability: The guideline requires the annotators to conduct a"comprehensive interrogation of the quantifiable parameters",considering "numerous facets and intricate variables", and also take into account "statistical significance thresholds" and"probability distributions". It's quite strict and complex, leaving little room for adaptability depending on the experimental intervention being evaluated. The rigidness of these instructions can make it challenging for evaluators to apply them across a variety of scenarios or different types of interventions.\\~Lack of Rating Scale: The guideline suggests that the final judgement should be a"synthesis of these insights, crystallized into a ranking". However,it doesn't provide any clear definition or structure for this ranking system.Without knowing how many points are on the scale or what each point represents,annotators might interpret the ranking system differently, leading to inconsistency in evaluations.~ ~~\end{tabular}                                                                                                                                                                                                                                                \\
\cellcolor[rgb]{0.9,0.9,0.9}{\textbf{\textit{NEGTIVE EXAMPLES}}}                                                                                                                                                                                                                                                                                                                                                                                                                                                                                                                                                                                                                                                                                                                                                                                                                                                                                                                                                                                                                                                                                                                                                                                                                                                                                                                                                                                                                                                                                                                                                                                                                                                                                                                                            \\
\textbf{Guideline:} This task aims to evaluate the machine translation quality of two different models. You will be given one article and two corresponding translations from the two models in a random order. Your task is to evaluate the quality of the two translation and determine which one you prefer.                                                                                                                                                                                                                                                                                                                                                                                                                                                                                                                                                                                                                                                                                                                                                                                                                                                                                                                                                                                                                                                                                                                                                                                                                                                                                                                                                                                                                                                                                                  \\
\textbf{Label:} Definition Ambiguity, Assuming Prior Knowledge \textbf{\textcolor{red}~ \XSolidBrush}                                                                                                                                                                                                                                                                                                                                                                                                                                                                                                                                                                                                                                                                                                                                                                                                                                                                                                                                                                                                                                                                                                                                                                                                                                                                                                                                                                                                                                                                                                                                                                                                                                                                                                \\
\begin{tabular}[c]{@{}p{18cm}@{}}\textbf{Explanation: } \\The guideline does not contain "Definition Ambiguity" error since it has clearly stated the objective of the task. There's no ambiguity about what the annotators are supposed to do, which leaves no room for misunderstanding about how the task is to be carried out. However, it falls under "Lack of Rating Scale" defect because the guideline doesn't provide a specific rating scale or evaluation criteria that the annotators can use to objectively assess the translations. For instance, the guideline could instruct them to rate each translation on a scale of 1-5 for various aspects such as accuracy, fluency, and grammatical correctness, with clear descriptions of what each point on the scale signifies. Besides, the guideline does not has the defect of "Assuming Prior Knowledge". Evaluating the quality of a translation does not inherently require annotators has specialized knowledge about machine translation models, algorithms, or technical jargon. The expectation is that annotators can read and understand both the source and target languages, and are thus capable of judging the quality of the translation.\\\\
\end{tabular}                                                                                                                                                                                                                                                                                                                                                                                                                                                                                                                                                                        \\
\hline
\end{tabular}
}
\caption{Full instructions given to annotators of the vulnerability detection task in human evaluation guidelines. 
}
\label{table4}

\end{table*}

\section{Annotation Details}
\label{sec:anno_details}
The annotators we recruited are four college students with College English Test-6
certificates who are fluent in both English and Chinese languages, with Chinese as their mother tongue. There are 1 females and 3 males, with an average age of around 24. 
Then we conduct a training process.
Specifically, we conducted an online meeting for annotator training, covering the interpretation of annotation guidelines, explanations and examples of various guideline vulnerabilities, clarification of relevant considerations, and a Q\&A session. To confirm their proficiency, annotators underwent a pre-annotation test, and only those who passed were allowed to proceed with the formal annotation.
Specifically, 10 guidelines are randomly sampled with 5 in Authentic Guidelines and 5 in Synthetic Guidelines respectively. We annotated them first. Then, we calculated the accuracy of each participant based on our annotation. Higher accuracy indicates a more consistent understanding of our guidelines. Annotators who achieve at least 80\% accuracy are considered qualified to continue the annotation. 

We used Cohen's kappa\citep{cohen1960coefficient}  to measure the inter-rater reliability. Considering that 
each label is independent and there are diverse label combinations for multi-label classification task, we do not require that two annotators provide completely identical label sets for each guideline. Instead, we assess the agreement between the two annotators in terms of each label they assign. 
Specifically, let $n$ be the number of guidelines to be labeled by $A$ and $B$ two annotators. $g$ is the number of distinct vulnerability labels, and $f_{ij}$ denotes the frequency of the number of subjects with the $i_{th}$ categorical response for annotator $A$ and the $j_{th}$ categorical response for annotator $B$. The kappa agreement is then calculated as:
\begin{center}
$p_0=\frac{1}{n} \sum_{i=1}^g f_{i i} $, \vspace{3mm}

$p_e=\frac{1}{n^2} \sum_{i=1}^g f_{i+} f_{+i} $,  \vspace{3mm}

$\kappa=\frac{p_0-p_e}{1-p_e}$, \vspace{3mm}
\end{center}
where $f_{i+}$  is the total for the $i_{th}$ row $f_{+i}$ and is the total for the $i_{th}$ column in the frequency table.

\section{Prompt Template for Vulnerability Detection}
\label{sec:Prompt_Template}

The prompt templates used for vulnerability type detection on human evaluation guidelines are illustrated in Figure \ref{Fig2}. As previously mentioned, we formulate four types of prompt templates: Basic, VDesc, CoT-Basic and CoT-VDesc in both zero-shot and few-shot scenarios. 
The zero-shot template of Basic template comprises of a Requirement + Constraints + Guideline framework. Requirement specifies the task motivation and remains consistent across all prompts; Constraints emphasize the desired output format such as \textit{“ Only reply with the names of vulnerabilities or ‘None'”}; and Guideline represents the input data. On this basis, the template of VDesc further introduces the description of vulnerability types, 
which is expressed as Requirement + Description + Constraints + Guideline. We formulated the two prompt templates with the consideration that incorporating descriptions of vulnerability types can boost model effectiveness by offering more specific knowledge, yet simultaneously, it might also potentially disrupt the model's performance for introducing extra information.

Regarding the few-shot prompts, we expanded on the zero-shot method by incorporating seven pseudo-examples that encompass all vulnerability types except for “Others”, some of which include multiple vulnerability types, so as to facilitate more appropriate model reasoning.

Additionally, we explore the CoT prompting technique, which elicits complex multi-step reasoning through step-by-step answer examples. For zero-shot prompts, we incorporate CoT by simply incorporating the phrase “Let’s think step by step" before each answer, without supplying any examples \citep{kojima2022large}. 
It is worth noting that the phrase “Let's think step by step” is absent in the 7-shot prompt differing from zero-shot scenario. Instead, the phrase is integrated into the reasoning process of examples, through which we observed an improvement in performance. Inspired by \citet{wang2022self}, we integrate the results of each reasoning over three runs and select the most consistent answer as the final answer set. 





\begin{figure*}[ht]
    \centering
    \includegraphics[width=\linewidth]{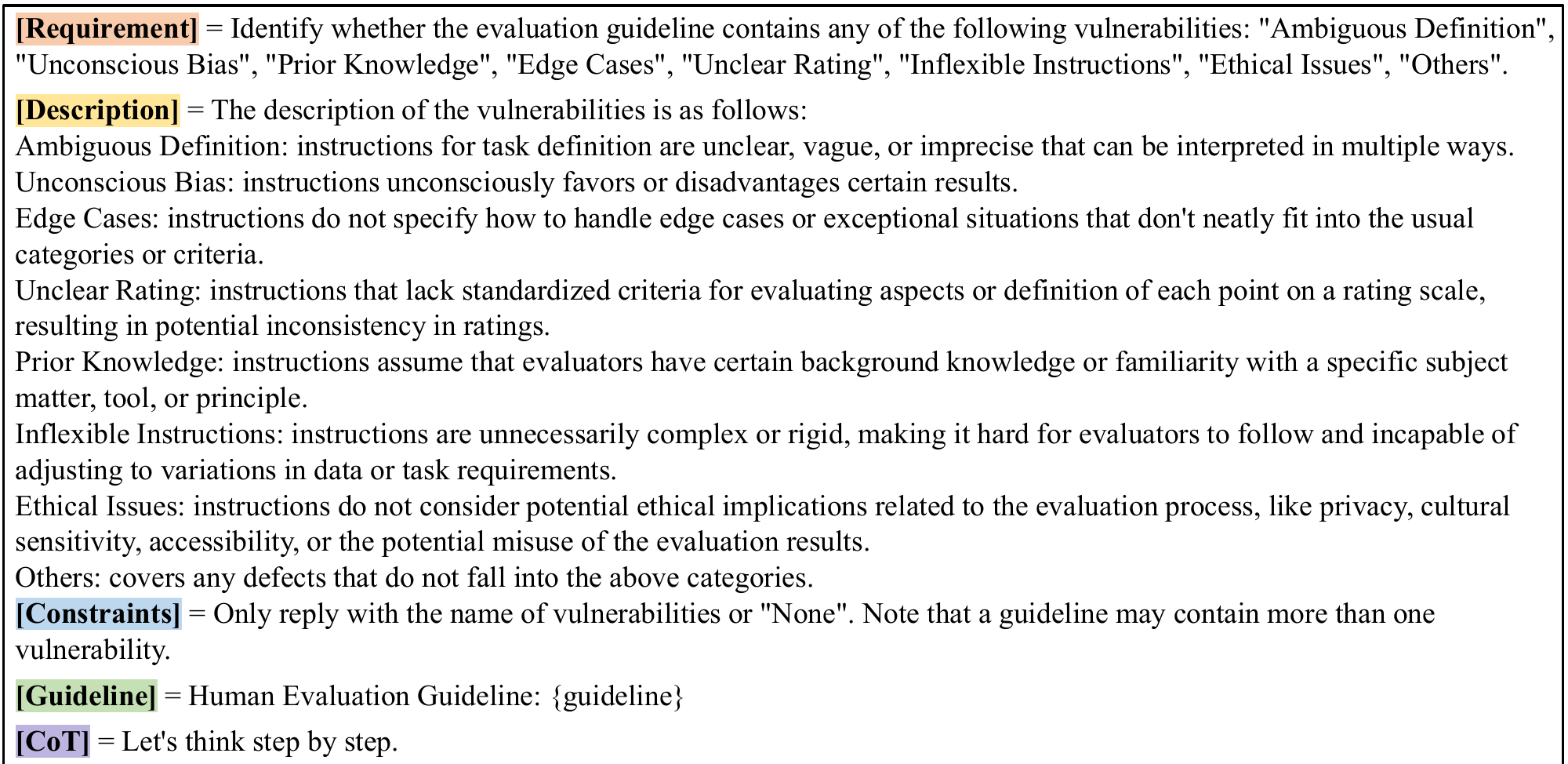}
    \includegraphics[width=\linewidth]{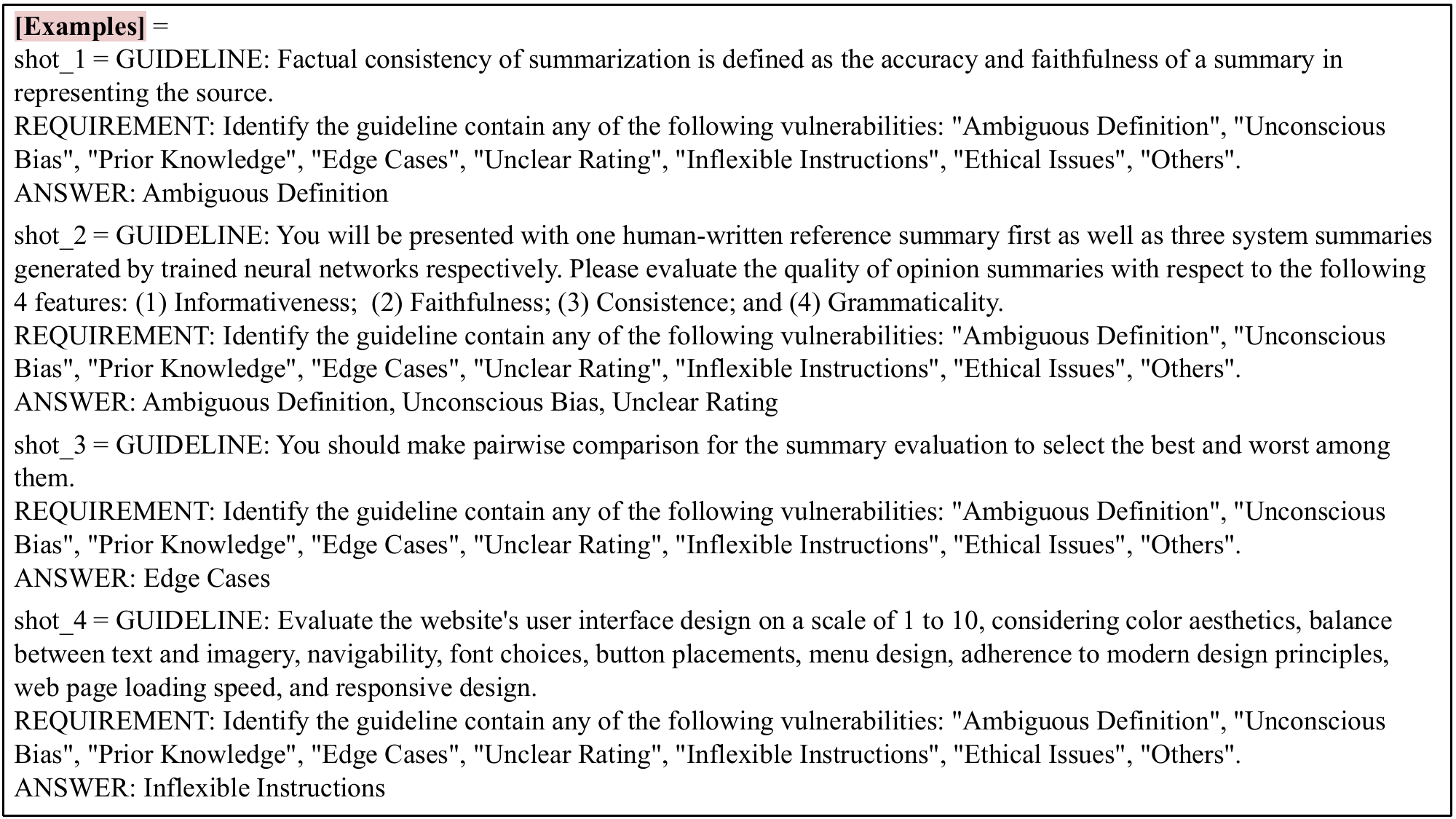}
    \includegraphics[width=\linewidth]{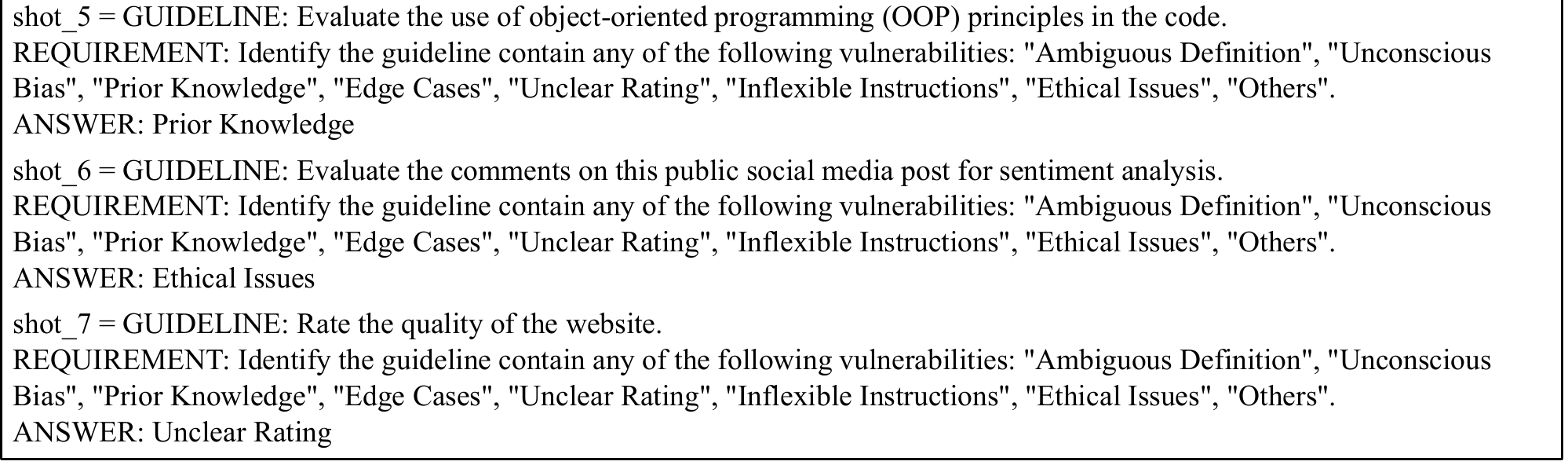}

    \end{figure*}

\begin{figure*}[ht]
    \centering
    \includegraphics[width=\linewidth]{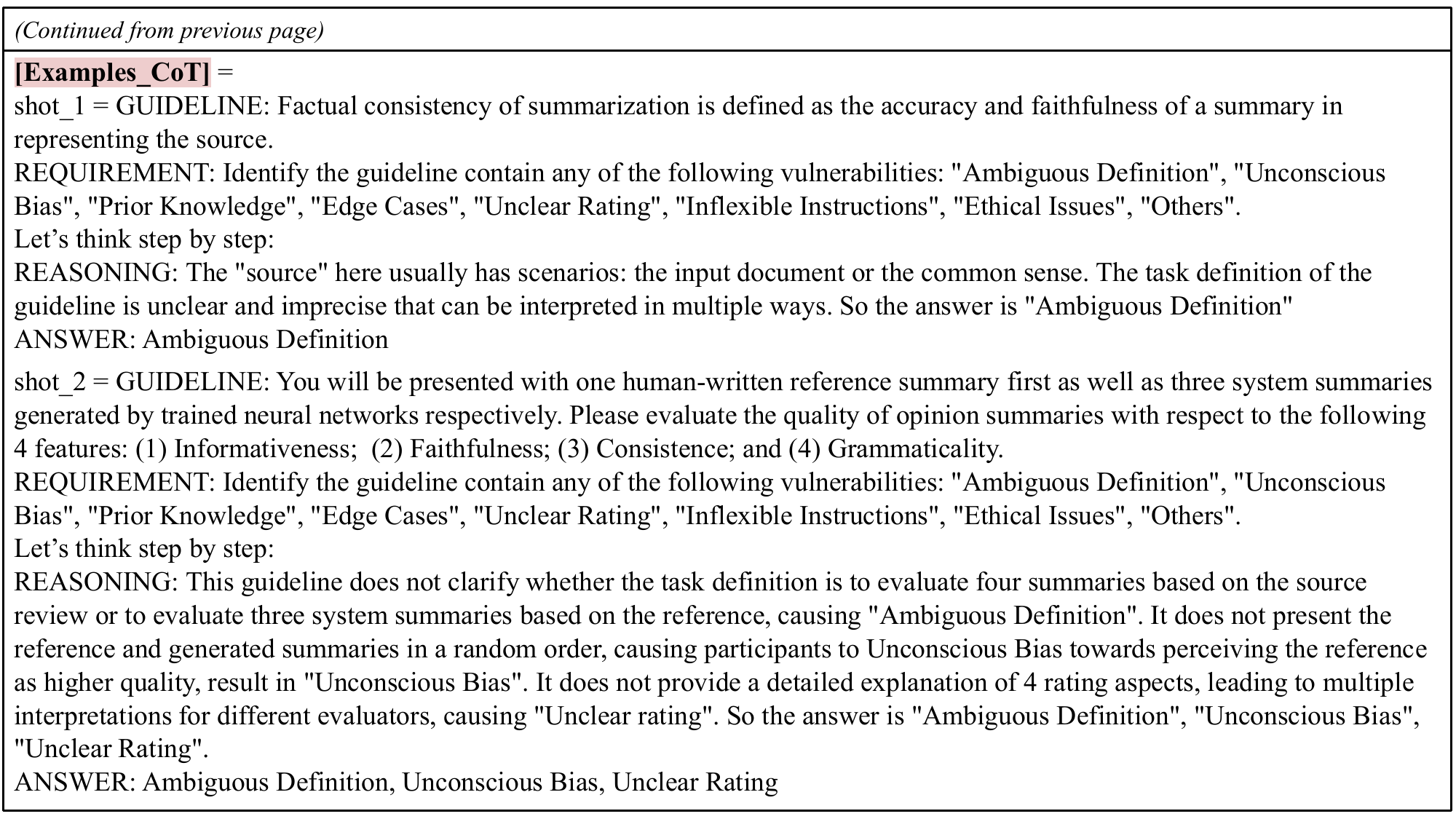}
    \includegraphics[width=\linewidth]{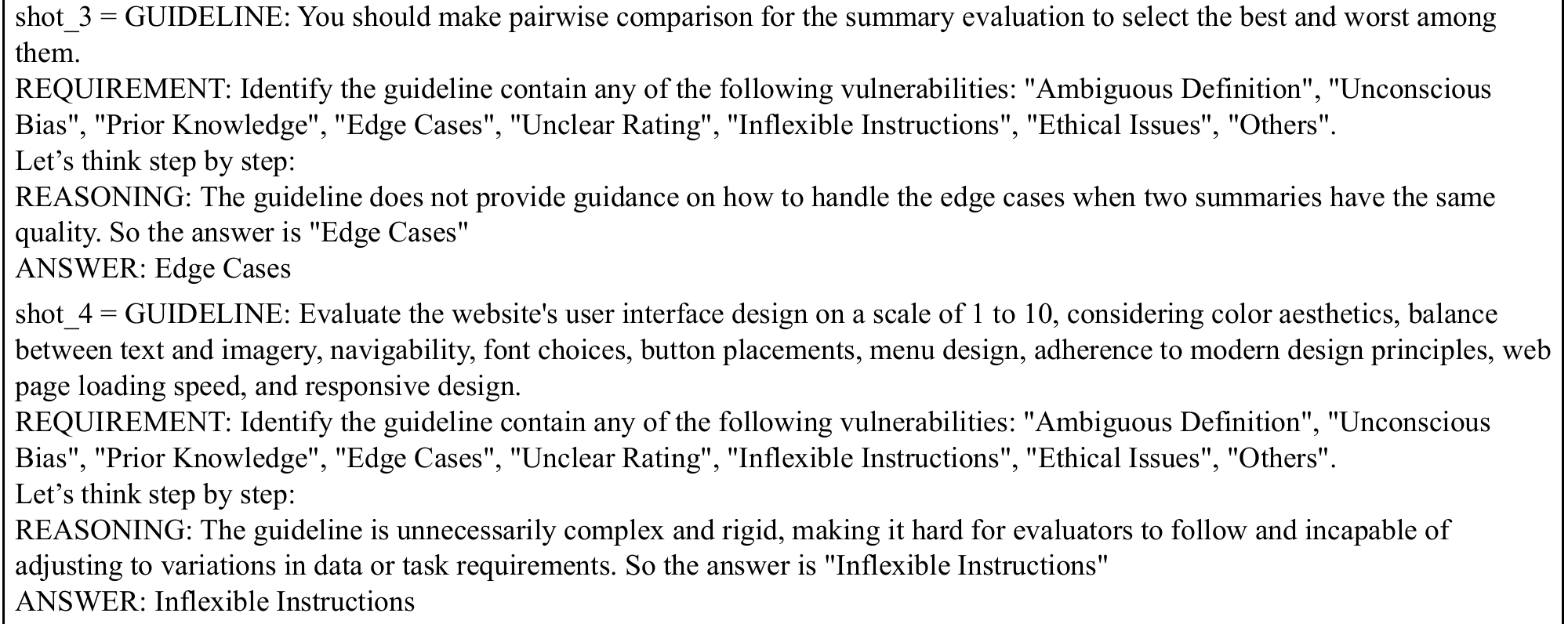}
    \includegraphics[width=\linewidth]{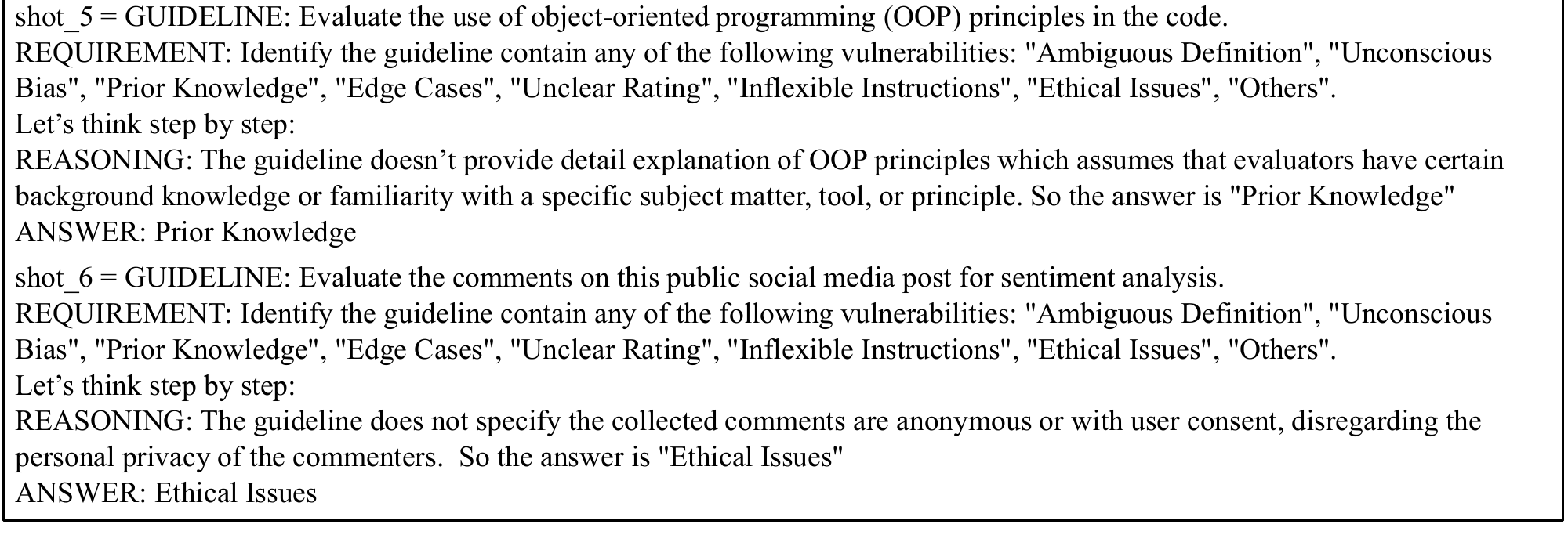}

    \end{figure*}

\begin{figure*}[ht]
    \centering
    \includegraphics[width=\linewidth]{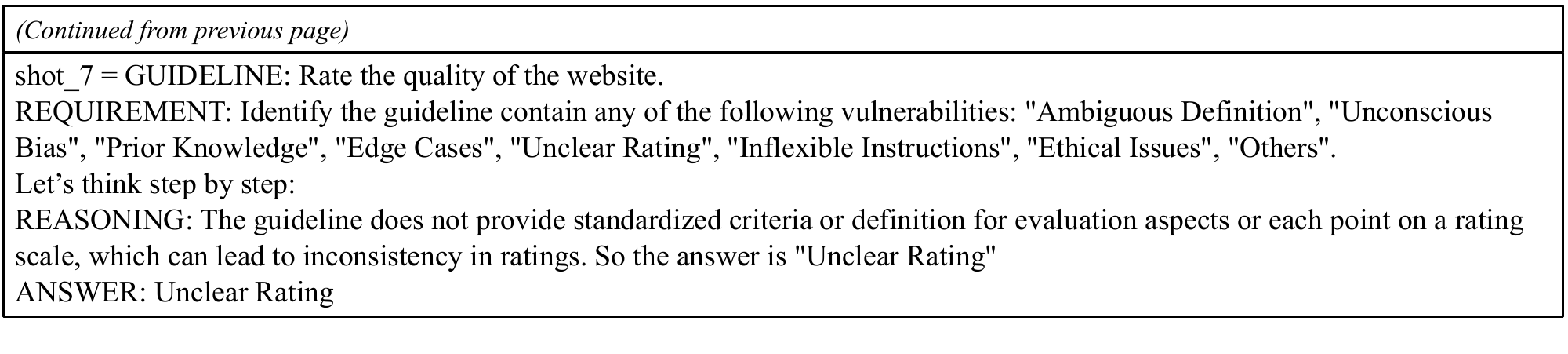}
    \includegraphics[width=\linewidth]{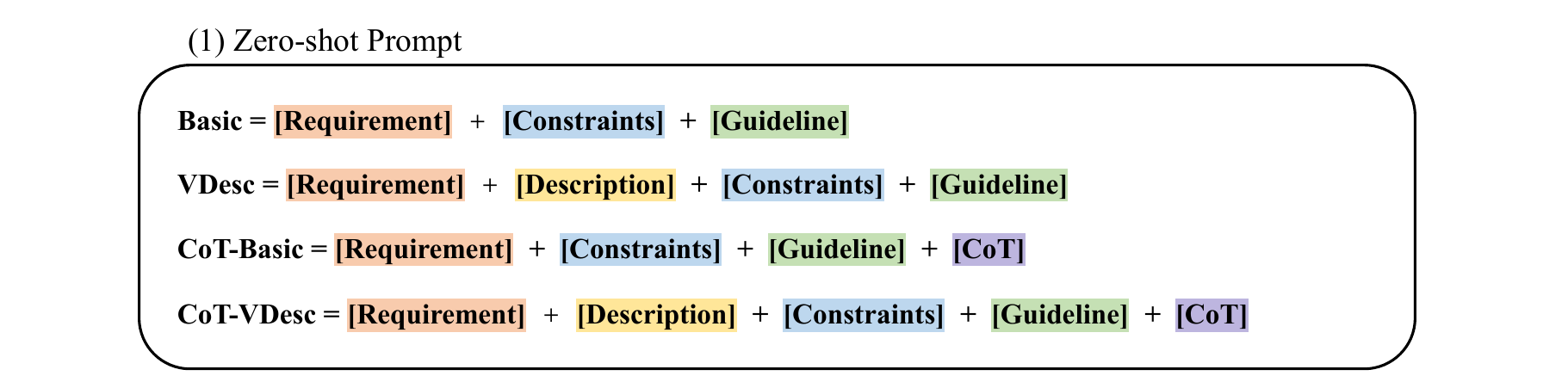}
    \includegraphics[width=\linewidth]{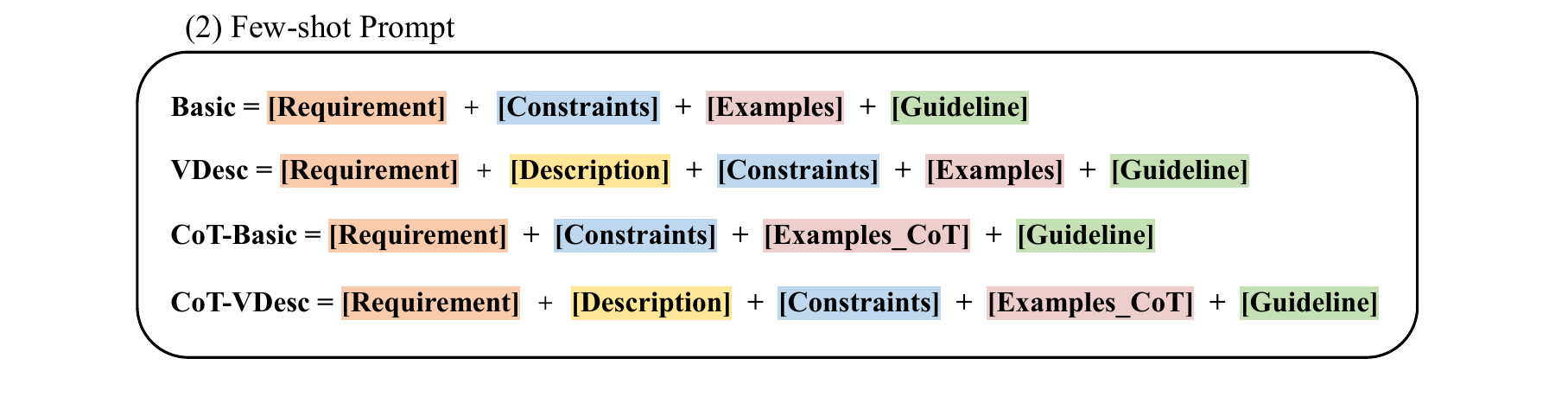}
    \caption{Full prompts containing Basic, VDesc and CoT used for vulnerability type detection.
    }
    \label{Fig2}
    \end{figure*}

\section{Hyper-parameters}
\label{sec:hyper-para}
\begin{table}[]
\centering
\renewcommand{\arraystretch}{1.1}
\begin{tabular}{cc}
\hline
\toprule
\multicolumn{2}{c}{\textbf{TEXT-DAVINCI-003}} \\ \cmidrule(r){1-2}
\multicolumn{1}{l|}{Max Tokens} & 1000 \\
\multicolumn{1}{l|}{Temperature} & 0 \\ \cmidrule(r){1-2}
\multicolumn{2}{c}{\textbf{GPT-3.5-Turbo}} \\ \cmidrule(r){1-2}
\multicolumn{1}{l|}{Max Tokens} & 1000 \\
\multicolumn{1}{l|}{Temperature} & 0 \\ \hline
\toprule
\end{tabular}
\caption{Hyper-parameters for TEXT-DAVINCI-003 and GPT-3.5-Turbo models.}
\label{tab:table12}
\end{table}

\begin{table}[ht]
\centering
\renewcommand{\arraystretch}{1.1}
\resizebox{\linewidth}{!}{
\begin{tabular}{l|l} 
\hline
\toprule
\multicolumn{2}{c}{\textbf{BERT \& XLNet \& ALBERT}} \\ \cmidrule(r){1-2}
Implementation Library   & Transformers \citep{devlin-etal-2019-bert}  \\
Computing Infrastructure & 12GB GeForce GTX 1080 Ti           \\
Max Seq Length           & 256                                \\
Optimizer                & AdamW                              \\
Optimizer Params:        &  $\beta =(0.9,0.999), \epsilon=1e-6$   \\
Learning rate            & 2e-5                               \\
Loss function            & BCELoss                               \\
Weight Decay             & 0.01                               \\
Maximum Gradient Norm    & 1.0                                \\
Batch size               & 8                                  \\
Epochs                   & 6                                  \\
\hline
\toprule
\end{tabular}
}
\caption{ Hyper-parameters for BERT, XLNet and ALBERT models.}
\label{table13}
\end{table}
For TEXT-DAVINCI-003\footnote{
https://platform.openai.com/docs/models/gpt-3-5} and GPT-3.5-Turbo\footnote{
https://platform.openai.com/docs/models/gpt-3-5}, the specific hyper-parameters can be found in Table \ref{tab:table12}.
For the baselines, we utilize Huggingface\footnote{https://huggingface.co/models} implementations for all the deep pretrained models and the specific hyper-parameters can be found in Table \ref{table13}. We initially explored two approaches: one is to treat the multi-label classification task as a seq2seq problem, and generate a variable-length label sequence for a given text sequence. The other is to consider each neuron to be a binary classifier since the predictions for each category are independent in multi-label classification task, essentially forming a binary classification task for each label. We ultimately chose the second approach due to the limited training dataset in this task (less than 500 samples) and the increased data requirements of complex sequence models.
We selected BCELoss as the loss function, which is commonly used in binary classification tasks. It calculates the individual losses for each label, quantifying the model's performance in terms of the difference between its predictions and the actual labels for each vulnerability type. Besides, we utilized the sigmoid activation function in the fully connected layers.

\section{Results of Each Vulnerability Type}
\label{sec:results_of_each_type}
Table \ref{tab11}, \ref{tab9} and \ref{tab10} 
report the experimental results of each vulnerability type (including “None”) of pre-trained baselines, 
TEXT-DAVINCI-003 as well as GPT-3.5-Turbo 
respectively. Please note that the overall accuracy and AUC in Table \ref{tab:tab6} are the averages across eight types of vulnerabilities, which may vary with the inclusion of “None”.
We can observe that the model's capacity to detect different vulnerability types exhibits some variation, showing a trend in both LLMs and the Baseline, where more frequently occurring vulnerability types yield lower results. For LLMs, this is reasonable as they tend to output “None”, as mentioned above, making them prone to misidentify more guidelines containing high-frequency vulnerability types as positive. Nevertheless, for the Baseline, which has undergone supervised training beforehand, we speculate this could be attributed to the limited size of the dataset, resulting in the models not having acquired robust capabilities yet. Additionally, the accuracy of “None” reaches highest at 0.79 in TEXT-DAVINCI-003 CoT-VDesc, which can be considered as an indicator of the reliability of the human evaluation guideline.

\begin{table}[ht]
\centering
\small
\renewcommand{\arraystretch}{1.1}
\resizebox{\linewidth}{!}{
\begin{tabular}{lllllll}
\hline
\toprule
\multicolumn{1}{c}{\textbf{Vulnerability   Type}} & \begin{tabular}[c]{@{}l@{}}Macro-\\ P\end{tabular} & \begin{tabular}[c]{@{}l@{}}Macro-\\ R\end{tabular} & \begin{tabular}[c]{@{}l@{}}Macro-\\ F1\end{tabular} & ACC & AUC & \begin{tabular}[c]{@{}l@{}}Hamming\\ Loss↓\end{tabular} \\ \hline
\multicolumn{7}{c}{\cellcolor[rgb]{0.9,0.9,0.9}{\textbf{BERT}}} \\ \cmidrule(r){1-7}
\multicolumn{1}{l|}{Ambiguous Definition} & 0.43 & 0.49 & 0.41 & 0.51 & 0.50 & 0.49 \\
\multicolumn{1}{l|}{Unconscious Bias} & 0.53 & 0.50 & 0.50 & 0.89 & 0.54 & 0.11 \\
\multicolumn{1}{l|}{Prior Knowledge} & 0.43 & 0.46 & 0.45 & 0.77 & 0.50 & 0.23 \\
\multicolumn{1}{l|}{Edge Cases} & 0.44 & 0.47 & 0.45 & 0.77 & 0.51 & 0.23 \\
\multicolumn{1}{l|}{Unclear Rating} & 0.65 & 0.60 & 0.57 & 0.69 & 0.59 & 0.31 \\
\multicolumn{1}{l|}{Inflexible Instructions} & 0.49 & 0.46 & 0.47 & 0.90 & 0.48 & 0.10 \\
\multicolumn{1}{l|}{Ethical Issues} & 0.47 & 0.47 & 0.47 & 0.86 & 0.53 & 0.14 \\
\multicolumn{1}{l|}{Others} & 0.49 & 0.47 & 0.48 & 0.91 & 0.48 & 0.09 \\
\multicolumn{1}{l|}{None} & 0.48 & 0.49 & 0.45 & 0.76 & 0.51 & 0.19 \\ \cmidrule(r){1-7}
\multicolumn{7}{c}{\cellcolor[rgb]{0.9,0.9,0.9}{\textbf{XLNET}}} \\ \cmidrule(r){1-7}
\multicolumn{1}{l|}{Ambiguous Definition} & 0.39 & 0.50 & 0.41 & 0.52 & 0.51 & 0.48 \\
\multicolumn{1}{l|}{Unconscious Bias} & 0.48 & 0.46 & 0.47 & 0.89 & 0.50 & 0.11 \\
\multicolumn{1}{l|}{Prior Knowledge} & 0.44 & 0.47 & 0.45 & 0.78 & 0.51 & 0.22 \\
\multicolumn{1}{l|}{Edge Cases} & 0.44 & 0.47 & 0.45 & 0.77 & 0.51 & 0.23 \\
\multicolumn{1}{l|}{Unclear Rating} & 0.55 & 0.58 & 0.56 & 0.65 & 0.56 & 0.35 \\
\multicolumn{1}{l|}{Inflexible Instructions} & 0.49 & 0.46 & 0.47 & 0.90 & 0.48 & 0.10 \\
\multicolumn{1}{l|}{Ethical Issues} & 0.49 & 0.49 & 0.49 & 0.87 & 0.54 & 0.13 \\
\multicolumn{1}{l|}{Others} & 0.49 & 0.47 & 0.48 & 0.91 & 0.48 & 0.09 \\
\multicolumn{1}{l|}{None} & 0.46 & 0.47 & 0.47 & 0.77 & 0.55 & 0.14 \\ \cmidrule(r){1-7}
\multicolumn{7}{c}{\cellcolor[rgb]{0.9,0.9,0.9}{\textbf{ALBERT}}} \\ \cmidrule(r){1-7}
\multicolumn{1}{l|}{Ambiguous Definition} & 0.39 & 0.50 & 0.41 & 0.52 & 0.51 & 0.48 \\
\multicolumn{1}{l|}{Unconscious Bias} & 0.48 & 0.47 & 0.47 & 0.90 & 0.51 & 0.10 \\
\multicolumn{1}{l|}{Prior Knowledge} & 0.42 & 0.47 & 0.44 & 0.80 & 0.51 & 0.20 \\
\multicolumn{1}{l|}{Edge Cases} & 0.45 & 0.48 & 0.46 & 0.79 & 0.53 & 0.21 \\
\multicolumn{1}{l|}{Unclear Rating} & 0.58 & 0.53 & 0.49 & 0.64 & 0.52 & 0.36 \\
\multicolumn{1}{l|}{Inflexible Instructions} & 0.49 & 0.48 & 0.48 & 0.93 & 0.49 & 0.07 \\
\multicolumn{1}{l|}{Ethical Issues} & 0.46 & 0.47 & 0.47 & 0.88 & 0.53 & 0.12 \\
\multicolumn{1}{l|}{Others} & 0.49 & 0.48 & 0.48 & 0.94 & 0.50 & 0.06 \\
\multicolumn{1}{l|}{None} & 0.47 & 0.48 & 0.46 & 0.77 & 0.50 & 0.19 \\ \hline
\toprule

\end{tabular}
}
\caption{Baseline results of each vulnerability type detection.}
\label{tab11}
\end{table}

\begin{table*}[t]
\centering
\small
\renewcommand{\arraystretch}{1.1}
\resizebox{\linewidth}{!}{
\begin{tabular}{lp{0.8cm}p{0.8cm}p{0.8cm}p{0.8cm}p{0.8cm}llp{0.8cm}p{0.8cm}p{0.8cm}p{0.8cm}p{0.8cm}p{0.8cm}}
\hline
\toprule
\multicolumn{1}{c|}{\multirow{2}{*}{\textbf{Vulnerability   Type}}} & \multicolumn{6}{c}{\textbf{Zero-shot}} &  & \multicolumn{6}{c}{\textbf{Few-shot}} \\ \cmidrule(r){2-7} \cmidrule(r){9-14} 
\multicolumn{1}{c|}{} & \begin{tabular}[c]{@{}l@{}}Macro-\\ P\end{tabular} & \begin{tabular}[c]{@{}l@{}}Macro-\\ R\end{tabular} & \begin{tabular}[c]{@{}l@{}}Macro-\\ F1\end{tabular} & ACC & AUC & \begin{tabular}[c]{@{}l@{}}Hamming\\ Loss↓\end{tabular} &  & \begin{tabular}[c]{@{}l@{}}Macro-\\ P\end{tabular} & \begin{tabular}[c]{@{}l@{}}Macro-\\ R\end{tabular} & \begin{tabular}[c]{@{}l@{}}Macro-\\ F1\end{tabular} & ACC & AUC & \begin{tabular}[c]{@{}l@{}}Hamming\\ Loss↓\end{tabular} \\ \cmidrule(r){1-14}
\multicolumn{14}{c}{\cellcolor[rgb]{0.9,0.9,0.9}{\textbf{Basic}}} \\ \hline
\multicolumn{1}{l|}{Ambiguous Definition} & 0.42 & 0.49 & 0.41 & 0.66 & 0.49 & 0.34 &  & 0.61 & 0.56 & 0.53 & 0.70 & 0.56 & 0.30 \\
\multicolumn{1}{l|}{Unconscious Bias} & 0.50 & 0.51 & 0.43 & 0.65 & 0.51 & 0.35 &  & 0.52 & 0.50 & 0.41 & 0.60 & 0.50 & 0.40 \\
\multicolumn{1}{l|}{Prior Knowledge} & 0.44 & 0.50 & 0.47 & 0.88 & 0.50 & 0.12 &  & 0.48 & 0.48 & 0.47 & 0.82 & 0.48 & 0.18 \\
\multicolumn{1}{l|}{Edge Cases} & 0.49 & 0.47 & 0.47 & 0.69 & 0.41 & 0.31 &  & 0.53 & 0.52 & 0.52 & 0.88 & 0.55 & 0.12 \\
\multicolumn{1}{l|}{Unclear Rating} & 0.43 & 0.42 & 0.40 & 0.41 & 0.42 & 0.59 &  & 0.51 & 0.50 & 0.50 & 0.53 & 0.51 & 0.47 \\
\multicolumn{1}{l|}{Inflexible Instructions} & 0.51 & 0.64 & 0.40 & 0.57 & 0.64 & 0.43 &  & 0.49 & 0.54 & 0.47 & 0.77 & 0.54 & 0.23 \\
\multicolumn{1}{l|}{Ethical Issues} & 0.49 & 0.48 & 0.48 & 0.87 & 0.48 & 0.13 &  & 0.57 & 0.62 & 0.56 & 0.85 & 0.62 & 0.15 \\
\multicolumn{1}{l|}{Others} & 0.48 & 0.49 & 0.49 & 0.95 & 0.49 & 0.05 &  & 0.55 & 0.57 & 0.57 & 0.97 & 0.57 & 0.03 \\
\multicolumn{1}{l|}{None} & 0.48 & 0.51 & 0.45 & 0.72 & 0.43 & 0.23 &  & 0.52 & 0.54 & 0.48 & 0.73 & 0.55 & 0.17 \\ \cmidrule(r){1-14}
\multicolumn{14}{c}{\cellcolor[rgb]{0.9,0.9,0.9}{\textbf{Vdesc}}} \\ \hline
\multicolumn{1}{l|}{Ambiguous Definition} & 0.34 & 0.50 & 0.40 & 0.67 & 0.50 & 0.33 &  & 0.55 & 0.58 & 0.48 & 0.67 & 0.55 & 0.33 \\
\multicolumn{1}{l|}{Unconscious Bias} & 0.48 & 0.48 & 0.48 & 0.90 & 0.48 & 0.07 &  & 0.50 & 0.49 & 0.46 & 0.78 & 0.53 & 0.22 \\
\multicolumn{1}{l|}{Prior Knowledge} & 0.44 & 0.50 & 0.47 & 0.80 & 0.50 & 0.12 &  & 0.49 & 0.48 & 0.48 & 0.84 & 0.54 & 0.16 \\
\multicolumn{1}{l|}{Edge Cases} & 0.44 & 0.47 & 0.45 & 0.77 & 0.54 & 0.17 &  & 0.53 & 0.56 & 0.51 & 0.88 & 0.61 & 0.12 \\
\multicolumn{1}{l|}{Unclear Rating} & 0.51 & 0.51 & 0.48 & 0.64 & 0.51 & 0.36 &  & 0.52 & 0.52 & 0.43 & 0.43 & 0.56 & 0.57 \\
\multicolumn{1}{l|}{Inflexible Instructions} & 0.50 & 0.50 & 0.48 & 0.81 & 0.50 & 0.16 &  & 0.59 & 0.57 & 0.47 & 0.73 & 0.65 & 0.27 \\
\multicolumn{1}{l|}{Ethical Issues} & 0.48 & 0.48 & 0.48 & 0.87 & 0.48 & 0.08 &  & 0.52 & 0.52 & 0.51 & 0.86 & 0.54 & 0.14 \\
\multicolumn{1}{l|}{Others} & 0.48 & 0.49 & 0.49 & 0.88 & 0.49 & 0.04 &  & 0.55 & 0.60 & 0.57 & 0.97 & 0.58 & 0.03 \\
\multicolumn{1}{l|}{None} & 0.48 & 0.49 & 0.49 & 0.75 & 0.57 & 0.15 &  & 0.53 & 0.53 & 0.50 & 0.74 & 0.56 & 0.18 \\ \cmidrule(r){1-14}
\multicolumn{14}{c}{\cellcolor[rgb]{0.9,0.9,0.9}{\textbf{CoT-Basic}}} \\ \cmidrule(r){1-14}
\multicolumn{1}{l|}{Ambiguous Definition} & 0.46 & 0.47 & 0.46 & 0.57 & 0.47 & 0.43 &  & 0.52 & 0.52 & 0.51 & 0.54 & 0.62 & 0.46 \\
\multicolumn{1}{l|}{Unconscious Bias} & 0.51 & 0.60 & 0.29 & 0.35 & 0.60 & 0.65 &  & 0.53 & 0.53 & 0.53 & 0.93 & 0.67 & 0.07 \\
\multicolumn{1}{l|}{Prior Knowledge} & 0.47 & 0.50 & 0.48 & 0.86 & 0.50 & 0.14 &  & 0.44 & 0.50 & 0.47 & 0.88 & 0.59 & 0.12 \\
\multicolumn{1}{l|}{Edge Cases} & 0.53 & 0.57 & 0.47 & 0.57 & 0.57 & 0.43 &  & 0.53 & 0.54 & 0.53 & 0.76 & 0.65 & 0.24 \\
\multicolumn{1}{l|}{Unclear Rating} & 0.45 & 0.48 & 0.29 & 0.33 & 0.60 & 0.67 &  & 0.44 & 0.44 & 0.37 & 0.37 & 0.58 & 0.63 \\
\multicolumn{1}{l|}{Inflexible Instructions} & 0.51 & 0.56 & 0.24 & 0.29 & 0.56 & 0.71 &  & 0.55 & 0.53 & 0.54 & 0.97 & 0.64 & 0.03 \\
\multicolumn{1}{l|}{Ethical Issues} & 0.49 & 0.47 & 0.44 & 0.73 & 0.47 & 0.27 &  & 0.48 & 0.49 & 0.49 & 0.95 & 0.57 & 0.05 \\
\multicolumn{1}{l|}{Others} & 0.48 & 0.46 & 0.47 & 0.89 & 0.45 & 0.11 &  & 0.48 & 0.50 & 0.49 & 0.97 & 0.62 & 0.03 \\
\multicolumn{1}{l|}{None} & 0.50 & 0.51 & 0.41 & 0.61 & 0.64 & 0.38 &  & 0.47 & 0.50 & 0.48 & 0.75 & 0.64 & 0.14 \\ \cmidrule(r){1-14}
\multicolumn{14}{c}{\cellcolor[rgb]{0.9,0.9,0.9}{\textbf{CoT-Vdesc}}} \\ \cmidrule(r){1-14}
\multicolumn{1}{l|}{Ambiguous Definition} & 0.47 & 0.47 & 0.47 & 0.55 & 0.52 & 0.45 &  & 0.60 & 0.60 & 0.60 & 0.67 & 0.68 & 0.34 \\
\multicolumn{1}{l|}{Unconscious Bias} & 0.54 & 0.61 & 0.51 & 0.81 & 0.61 & 0.19 &  & 0.54 & 0.53 & 0.53 & 0.89 & 0.59 & 0.11 \\
\multicolumn{1}{l|}{Prior Knowledge} & 0.46 & 0.49 & 0.47 & 0.84 & 0.51 & 0.16 &  & 0.68 & 0.51 & 0.49 & 0.86 & 0.78 & 0.14 \\
\multicolumn{1}{l|}{Edge Cases} & 0.52 & 0.55 & 0.50 & 0.66 & 0.54 & 0.34 &  & 0.65 & 0.67 & 0.64 & 0.81 & 0.71 & 0.19 \\
\multicolumn{1}{l|}{Unclear Rating} & 0.49 & 0.48 & 0.39 & 0.39 & 0.59 & 0.61 &  & 0.65 & 0.65 & 0.58 & 0.58 & 0.69 & 0.42 \\
\multicolumn{1}{l|}{Inflexible Instructions} & 0.51 & 0.55 & 0.23 & 0.27 & 0.55 & 0.73 &  & 0.64 & 0.65 & 0.65 & 0.97 & 0.74 & 0.07 \\
\multicolumn{1}{l|}{Ethical Issues} & 0.50 & 0.46 & 0.42 & 0.63 & 0.51 & 0.37 &  & 0.56 & 0.52 & 0.53 & 0.93 & 0.62 & 0.07 \\
\multicolumn{1}{l|}{Others} & 0.48 & 0.50 & 0.49 & 0.96 & 0.50 & 0.04 &  & 0.48 & 0.50 & 0.49 & 0.96 & 0.63 & 0.04 \\
\multicolumn{1}{l|}{None} & 0.51 & 0.50 & 0.43 & 0.66 & 0.68 & 0.30 &  & 0.58 & 0.57 & 0.54 & 0.79 & 0.69 & 0.10 \\ \hline
\toprule

\end{tabular}
}
\caption{TEXT-DAVINCI-003 results of each vulnerability type detection.}
\label{tab9}
\end{table*}

\begin{table*}[t]
\centering
\small
\renewcommand{\arraystretch}{1.1}
\resizebox{\linewidth}{!}{
\begin{tabular}{lp{0.8cm}p{0.8cm}p{0.8cm}p{0.8cm}p{0.8cm}llp{0.8cm}p{0.8cm}p{0.8cm}p{0.8cm}p{0.8cm}p{0.8cm}}
\hline
\toprule
\multicolumn{1}{c|}{\multirow{2}{*}{\textbf{Vulnerability   Type}}} & \multicolumn{6}{c}{\textbf{Zero-shot}} &  & \multicolumn{6}{c}{\textbf{Few-shot}} \\ \cmidrule(r){2-7} \cmidrule(r){9-14} 
\multicolumn{1}{c|}{} & \begin{tabular}[c]{@{}l@{}}Macro-\\ P\end{tabular} & \begin{tabular}[c]{@{}l@{}}Macro-\\ R\end{tabular} & \begin{tabular}[c]{@{}l@{}}Macro-\\ F1\end{tabular} & ACC & AUC & \begin{tabular}[c]{@{}l@{}}Hamming\\ Loss↓\end{tabular} &  & \begin{tabular}[c]{@{}l@{}}Macro-\\ P\end{tabular} & \begin{tabular}[c]{@{}l@{}}Macro-\\ R\end{tabular} & \begin{tabular}[c]{@{}l@{}}Macro-\\ F1\end{tabular} & ACC & AUC & \begin{tabular}[c]{@{}l@{}}Hamming\\ Loss↓\end{tabular} \\ \cmidrule(r){1-14}
\multicolumn{14}{c}{\cellcolor[rgb]{0.9,0.9,0.9}{\textbf{Basic}}} \\ \hline
\multicolumn{1}{l|}{Ambiguous Definition} & 0.47 & 0.48 & 0.46 & 0.55 & 0.52 & 0.45 &  & 0.57 & 0.57 & 0.57 & 0.60 & 0.57 & 0.40 \\
\multicolumn{1}{l|}{Unconscious Bias} & 0.50 & 0.51 & 0.34 & 0.44 & 0.54 & 0.56 &  & 0.52 & 0.55 & 0.50 & 0.77 & 0.55 & 0.23 \\
\multicolumn{1}{l|}{Prior Knowledge} & 0.49 & 0.56 & 0.50 & 0.75 & 0.58 & 0.25 &  & 0.51 & 0.50 & 0.49 & 0.80 & 0.51 & 0.20 \\
\multicolumn{1}{l|}{Edge Cases} & 0.53 & 0.58 & 0.47 & 0.55 & 0.62 & 0.45 &  & 0.42 & 0.50 & 0.46 & 0.85 & 0.50 & 0.15 \\
\multicolumn{1}{l|}{Unclear Rating} & 0.50 & 0.50 & 0.42 & 0.47 & 0.54 & 0.53 &  & 0.29 & 0.41 & 0.34 & 0.51 & 0.41 & 0.49 \\
\multicolumn{1}{l|}{Inflexible Instructions} & 0.53 & 0.78 & 0.41 & 0.56 & 0.57 & 0.44 &  & 0.56 & 0.92 & 0.58 & 0.92 & 0.58 & 0.08 \\
\multicolumn{1}{l|}{Ethical Issues} & 0.48 & 0.49 & 0.46 & 0.78 & 0.51 & 0.22 &  & 0.70 & 0.56 & 0.58 & 0.92 & 0.56 & 0.08 \\
\multicolumn{1}{l|}{Others} & 0.48 & 0.48 & 0.48 & 0.93 & 0.52 & 0.07 &  & 0.54 & 0.61 & 0.55 & 0.90 & 0.61 & 0.10 \\
\multicolumn{1}{l|}{None} & 0.51 & 0.53 & 0.44 & 0.65 & 0.62 & 0.33 &  & 0.50 & 0.57 & 0.50 & 0.76 & 0.45 & 0.17 \\ \cmidrule(r){1-14}
\multicolumn{14}{c}{\cellcolor[rgb]{0.9,0.9,0.9}{\textbf{Vdesc}}} \\ \hline
\multicolumn{1}{l|}{Ambiguous Definition} & 0.53 & 0.55 & 0.50 & 0.51 & 0.61 & 0.49 &  & 0.56 & 0.55 & 0.54 & 0.60 & 0.55 & 0.40 \\
\multicolumn{1}{l|}{Unconscious Bias} & 0.58 & 0.60 & 0.49 & 0.72 & 0.65 & 0.28 &  & 0.56 & 0.70 & 0.55 & 0.77 & 0.70 & 0.23 \\
\multicolumn{1}{l|}{Prior Knowledge} & 0.47 & 0.53 & 0.44 & 0.66 & 0.58 & 0.34 &  & 0.62 & 0.60 & 0.61 & 0.80 & 0.60 & 0.20 \\
\multicolumn{1}{l|}{Edge Cases} & 0.56 & 0.64 & 0.54 & 0.65 & 0.69 & 0.35 &  & 0.42 & 0.49 & 0.46 & 0.84 & 0.49 & 0.16 \\
\multicolumn{1}{l|}{Unclear Rating} & 0.55 & 0.54 & 0.47 & 0.48 & 0.59 & 0.52 &  & 0.45 & 0.48 & 0.43 & 0.56 & 0.48 & 0.44 \\
\multicolumn{1}{l|}{Inflexible Instructions} & 0.52 & 0.71 & 0.33 & 0.44 & 0.61 & 0.56 &  & 0.55 & 0.91 & 0.56 & 0.91 & 0.48 & 0.09 \\
\multicolumn{1}{l|}{Ethical Issues} & 0.51 & 0.56 & 0.45 & 0.69 & 0.61 & 0.31 &  & 0.47 & 0.49 & 0.48 & 0.91 & 0.49 & 0.09 \\
\multicolumn{1}{l|}{Others} & 0.48 & 0.49 & 0.48 & 0.94 & 0.54 & 0.06 &  & 0.52 & 0.59 & 0.52 & 0.87 & 0.60 & 0.13 \\
\multicolumn{1}{l|}{None} & 0.54 & 0.59 & 0.46 & 0.65 & 0.66 & 0.30 &  & 0.50 & 0.58 & 0.50 & 0.75 & 0.33 & 0.18 \\ \cmidrule(r){1-14}
\multicolumn{14}{c}{\cellcolor[rgb]{0.9,0.9,0.9}{\textbf{CoT-Basic}}} \\ \hline
\multicolumn{1}{l|}{Ambiguous Definition} & 0.58 & 0.58 & 0.58 & 0.64 & 0.58 & 0.36 &  & 0.55 & 0.55 & 0.54 & 0.59 & 0.55 & 0.41 \\
\multicolumn{1}{l|}{Unconscious Bias} & 0.50 & 0.50 & 0.45 & 0.68 & 0.52 & 0.32 &  & 0.53 & 0.60 & 0.51 & 0.73 & 0.58 & 0.27 \\
\multicolumn{1}{l|}{Prior Knowledge} & 0.50 & 0.49 & 0.47 & 0.63 & 0.54 & 0.37 &  & 0.44 & 0.48 & 0.46 & 0.77 & 0.49 & 0.23 \\
\multicolumn{1}{l|}{Edge Cases} & 0.57 & 0.61 & 0.57 & 0.73 & 0.69 & 0.27 &  & 0.62 & 0.69 & 0.64 & 0.81 & 0.69 & 0.19 \\
\multicolumn{1}{l|}{Unclear Rating} & 0.56 & 0.56 & 0.56 & 0.59 & 0.65 & 0.41 &  & 0.49 & 0.50 & 0.45 & 0.60 & 0.52 & 0.40 \\
\multicolumn{1}{l|}{Inflexible Instructions} & 0.53 & 0.81 & 0.50 & 0.79 & 0.66 & 0.21 &  & 0.50 & 0.50 & 0.50 & 0.97 & 0.63 & 0.03 \\
\multicolumn{1}{l|}{Ethical Issues} & 0.50 & 0.48 & 0.47 & 0.79 & 0.54 & 0.21 &  & 0.47 & 0.48 & 0.48 & 0.89 & 0.52 & 0.11 \\
\multicolumn{1}{l|}{Others} & 0.48 & 0.48 & 0.48 & 0.93 & 0.54 & 0.07 &  & 0.48 & 0.49 & 0.49 & 0.96 & 0.53 & 0.04 \\
\multicolumn{1}{l|}{None} & 0.51 & 0.56 & 0.49 & 0.71 & 0.69 & 0.26 &  & 0.49 & 0.54 & 0.49 & 0.74 & 0.60 & 0.17 \\ \cmidrule(r){1-14}
\multicolumn{14}{c}{\cellcolor[rgb]{0.9,0.9,0.9}\textbf{CoT-Vdesc}} \\ \cmidrule(r){1-14}
\multicolumn{1}{l|}{Ambiguous Definition} & 0.59 & 0.58 & 0.57 & 0.66 & 0.58 & 0.34 &  & 0.60 & 0.58 & 0.58 & 0.64 & 0.59 & 0.36 \\
\multicolumn{1}{l|}{Unconscious Bias} & 0.50 & 0.50 & 0.47 & 0.76 & 0.54 & 0.24 &  & 0.50 & 0.49 & 0.48 & 0.79 & 0.53 & 0.20 \\
\multicolumn{1}{l|}{Prior Knowledge} & 0.50 & 0.49 & 0.49 & 0.69 & 0.52 & 0.31 &  & 0.60 & 0.61 & 0.60 & 0.79 & 0.62 & 0.21 \\
\multicolumn{1}{l|}{Edge Cases} & 0.57 & 0.58 & 0.58 & 0.79 & 0.62 & 0.21 &  & 0.52 & 0.54 & 0.52 & 0.74 & 0.54 & 0.26 \\
\multicolumn{1}{l|}{Unclear Rating} & 0.53 & 0.53 & 0.52 & 0.58 & 0.58 & 0.42 &  & 0.60 & 0.55 & 0.51 & 0.65 & 0.55 & 0.36 \\
\multicolumn{1}{l|}{Inflexible Instructions} & 0.52 & 0.74 & 0.50 & 0.81 & 0.64 & 0.19 &  & 0.90 & 0.90 & 0.90 & 0.98 & 0.72 & 0.02 \\
\multicolumn{1}{l|}{Ethical Issues} & 0.51 & 0.54 & 0.50 & 0.81 & 0.57 & 0.19 &  & 0.58 & 0.63 & 0.59 & 0.87 & 0.63 & 0.13 \\
\multicolumn{1}{l|}{Others} & 0.48 & 0.49 & 0.49 & 0.95 & 0.52 & 0.05 &  & 0.48 & 0.50 & 0.49 & 0.97 & 0.56 & 0.04 \\
\multicolumn{1}{l|}{None} & 0.52 & 0.55 & 0.50 & 0.74 & 0.63 & 0.20 &  & 0.48 & 0.58 & 0.58 & 0.73 & 0.52 & 0.14 \\ \hline
\toprule

\end{tabular}
}
\caption{GPT-3.5-Turbo results of each vulnerability type detection.}
\label{tab10}
\end{table*}

\section{Principal for Reliable Human Evaluation Guideline}
\label{sec:principal_guideline}
A reliable human evaluation guideline is the beginning of reliable human evaluation.
We provide the principal for composing a reliable human evaluation guideline in Table \ref{table5}. For writing a reliable human evaluation guideline, researchers should provide explicit task definitions for raters and avoid biased instruction and prior knowledge assumptions. Moreover, the instruction should comprehensively cover a broad range of scenarios including the edge cases. Researchers should provide clear rating scale and criteria and make the instruction simple and flexible. Additionally, the potential ethical issues should be identified and addressed. It is highly recommended to attach examples and design a good user interface. Last but not least, remind annotators to be careful while carrying out their tasks to get more accurate evaluation results.

\begin{table*}[t]
\centering
\resizebox{\linewidth}{!}{
\begin{tabular}{|p{18cm}|} 
\hline

\multicolumn{1}{|c|}{\cellcolor[rgb]{0.9,0.9,0.9}{\large\textbf{Principal for Human Evaluation Guideline}}}                                                                                                                                                                                                                                                                                                                                                                                                               \\
\hline
\\
\textbf{(a) Explicit Task Definition: }Provide a concise and precise task description along with the evaluation aspects to preclude misinterpretation or confusion. Use clear language and avoid jargon or technical terms that may not be commonly known to all evaluators.~~                                                                                                                                                                                                          \\
\textbf{(b) Unbiased Instructions:} Ensure that all instructions and statements are free from any unconscious bias. Avoid favoring or disadvantaging certain results or approaches. Use neutral language and present the task in a fair and objective manner.~~                                                                                                                                                                                                                         \\
\textbf{(c) Avoiding Prior Knowledge Assumptions:} Provide sufficient explanations regarding the subject matter, tools, or principles involved. Avoid assuming that evaluators are equipped with specific background knowledge. A good guideline is to make the content easily comprehensible, even for non-expert annotators.~                                                                                                                                                         \\
\textbf{(d) Comprehensive Coverage:} Formulate instructions that cover a broad range of scenarios, incorporating edge cases and exceptional situations that may not fit neatly into predefined categories or criteria. Clearly specify how evaluators should handle such cases and provide guidance on making informed judgments.~ ~ ~ ~ ~ ~                                                                                                                                            \\
\textbf{(e) Clear Rating Scale and Criteria: }Define a rating scale and provide a clear explanation of the evaluation criteria for each point on the scale. Ensure that evaluators comprehensively grasp the meaning and expectations associated with each rating level so as to promote consistency in ratings and minimize potential confusion.~~                                                                                                                                     \\
\textbf{(f) Simplicity and Flexibility: }Keep the instructions straightforward and easy to understand. Avoid unnecessary complexity or rigid requirements that may make it difficult for evaluators to follow or adapt to variations in data and task requirements. Provide room for reasonable judgment and adaptability within the evaluation process.~~                                                                                                                              \\
\textbf{(g) Addressing Ethical Issues:} Identify and address potential ethical issues, including guidelines and safeguards, to ensure ethical considerations are upheld throughout the evaluation process. Consider privacy concerns, cultural sensitivities, accessibility requirements, and the potential misuse of evaluation results.~~                                                                                                                                             \\
\textbf{(h) Attach Examples:} It is highly recommended to list positive examples that contain the input presented to the worker or system, and the anticipated results, thus providing crowdworkers with a clearer comprehension of the task. Additionally, listing negative examples can effectively emphasize THINGS TO AVOID by supplying that should not be generated.~~                                                                                                            \\
\textbf{(i) Design a good user interface:} A good user interface provides a positive experience for annotators and the design needs to be tailored to the specific needs of the users. For non-expert crowdsourcing,~\textit{acquisition interfaces}~can be developed to facilitate the execution of crowdsourcing tasks. While for those running the crowdsourcing project,~\textit{management interfaces~}are required to monitor progress, assess quality, and manage annotators.~~  \\
\textbf{(j) Emphasize precautions: }In concluding the guidelines, remind annotators to be careful while carrying out their tasks and outline the annotation requirements, feedback mechanism, and quality assurance processes explicitly so that annotators can manage their time effectively and provide more accurate evaluations.~ ~ ~ ~ ~ ~ ~ ~~                                                                                                                                    \\
\\Adherence to these principles facilitates the creation of a human evaluation guideline that is clear, fair, inclusive, and capable of producing reliable and meaningful results.~ ~~   \\\\                                                                                                                                                                                                                                                                                                 \\
\hline
\end{tabular}
}
\caption{Principal for reliable human evaluation guideline. 
}
\label{table5}
\end{table*}

\end{document}